\documentclass[]{bytedance_seed}

\usepackage[toc,page,header]{appendix}
\usepackage{minitoc}
\usepackage{microtype}
\usepackage{graphicx}
\usepackage{subcaption}
\usepackage{booktabs} 
\usepackage{hyperref}
\usepackage{bm}

\usepackage{amsmath}
\usepackage{amssymb}
\usepackage{mathtools}
\usepackage{amsthm}
\theoremstyle{plain}

\theoremstyle{definition}

\theoremstyle{remark}

\usepackage{xspace}
\usepackage{wrapfig}
\usepackage{multirow}   
\usepackage[table,xcdraw]{xcolor} 
\usepackage{tabularx}      
\usepackage{colortbl}
\usepackage{float}
\usepackage{footnote}
\usepackage{stfloats}
\fnbelowfloat
\usepackage{afterpage}

\definecolor{myblue}{RGB}{45,91,168}
\definecolor{mygreen}{RGB}{0,153,229}
\newcommand{\method}{\textcolor{myblue}{WVM}\xspace}
\newcommand{\bench}{\textcolor{mygreen}{Suboptimal-Value-Bench}\xspace}

\newcommand{\figref}[1]{\hyperref[#1]{Fig.~\ref*{#1}}}
\newcommand{\tabref}[1]{\hyperref[#1]{Table~\ref*{#1}}}
\newcommand{\secref}[1]{\hyperref[#1]{Section~\ref*{#1}}}
\newcommand{\appref}[1]{\hyperref[#1]{Appendix~\ref*{#1}}}
\newcommand{\eqnref}[1]{\hyperref[#1]{Eq.~\ref*{#1}}}

\definecolor{SkyBlue}{HTML}{87CEEB}
\definecolor{cGVL}{HTML}{8D9E66}
\definecolor{cVLAC}{HTML}{66A1A1}
\definecolor{cRoboDopamine}{HTML}{66CC99}
\definecolor{cRoboReward}{HTML}{9999CC}
\definecolor{cRoboMeter}{HTML}{B30000}
\definecolor{cTopReward}{HTML}{B26666}

\title{World Value Models for Robotic Manipulation}

\author[1,2,3]{Zhihao Wang}
\author[1,3, \dagger]{Jianxiong Li}
\author[1, \S]{Yu Cui}
\author[3]{Yuan Gao}
\author[3]{\\Xianyuan Zhan} 
\author[2, \S]{Junzhi Yu} 
\author[1]{Xiao Ma} 

\affiliation[1]{ByteDance Seed}
\affiliation[2]{Peking University}
\affiliation[3]{Tsinghua University}

\contribution[\dagger]{Project Lead}
\contribution[\S]{Corresponding Author}
\date{\today}
\correspondence{\email{cuiyu.0627@bytedance.com}, \email{yujunzhi@pku.edu.cn}}
\abstract{
Generalist value models play a pivotal role in scaling robotic policy learning from large-scale, mixed-quality data.
Mathematically, accurate value estimation demands deep temporal understanding, requiring models to both ground the current belief using historical context and plan over future outcomes.
However, most existing robotic value models are built on Vision-Language Model (VLM) backbones that are pretrained primarily on static or temporally sparse visual observations, lacking the requisite temporal modeling capabilities for value estimation.
Unlike VLMs, world models naturally excel at temporal modeling and future planning, making them ideal foundations for learning generalizable value functions. Driven by this insight, we marry world models with value estimation to construct a new generalist robotic value model, \textbf{World Value Model} (\textbf{\method}), that offers accurate task progressions to assess data quality.
On standard benchmarks, \method delivers state-of-the-art (SOTA)  Value-Order Correlation (VOC) results.
Complementing standard evaluation suites that contains only expert data, we further introduce \textbf{\bench}, a multi-embodiment benchmark consisting of $800$ suboptimal trajectories with high-fidelity, human-labeled frame annotations. Our evaluations show that \method maintains its SOTA performance on \bench, establishing its robustness in handling both expert and suboptimal data.
When deployed for policy learning, \method improves manipulation performance across various policy extraction approaches in both simulated and real-world deployment, providing robust guidance for learning from mixed-quality data.
}

\checkdata[Project Page]{\href{https://zh1hao.wang/wvm}{\nolinkurl{zh1hao.wang/wvm}}}

\fancypagestyle{firststyle}{
    \fancyhead[R]{}
    \fancyhead[C]{}
    \fancyhead[L]{%
        \raisebox{-8mm}[0pt][0pt]{%
            \makebox[\headwidth][l]{%
                \includegraphics[width=53.3mm,height=6mm]{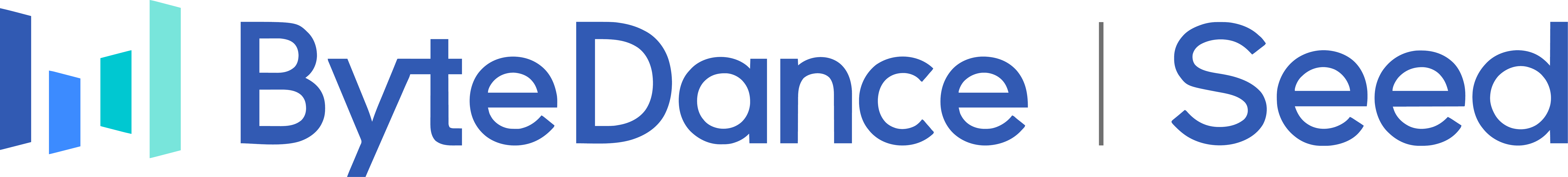}%
                \hfill
                \includegraphics[height=10mm]{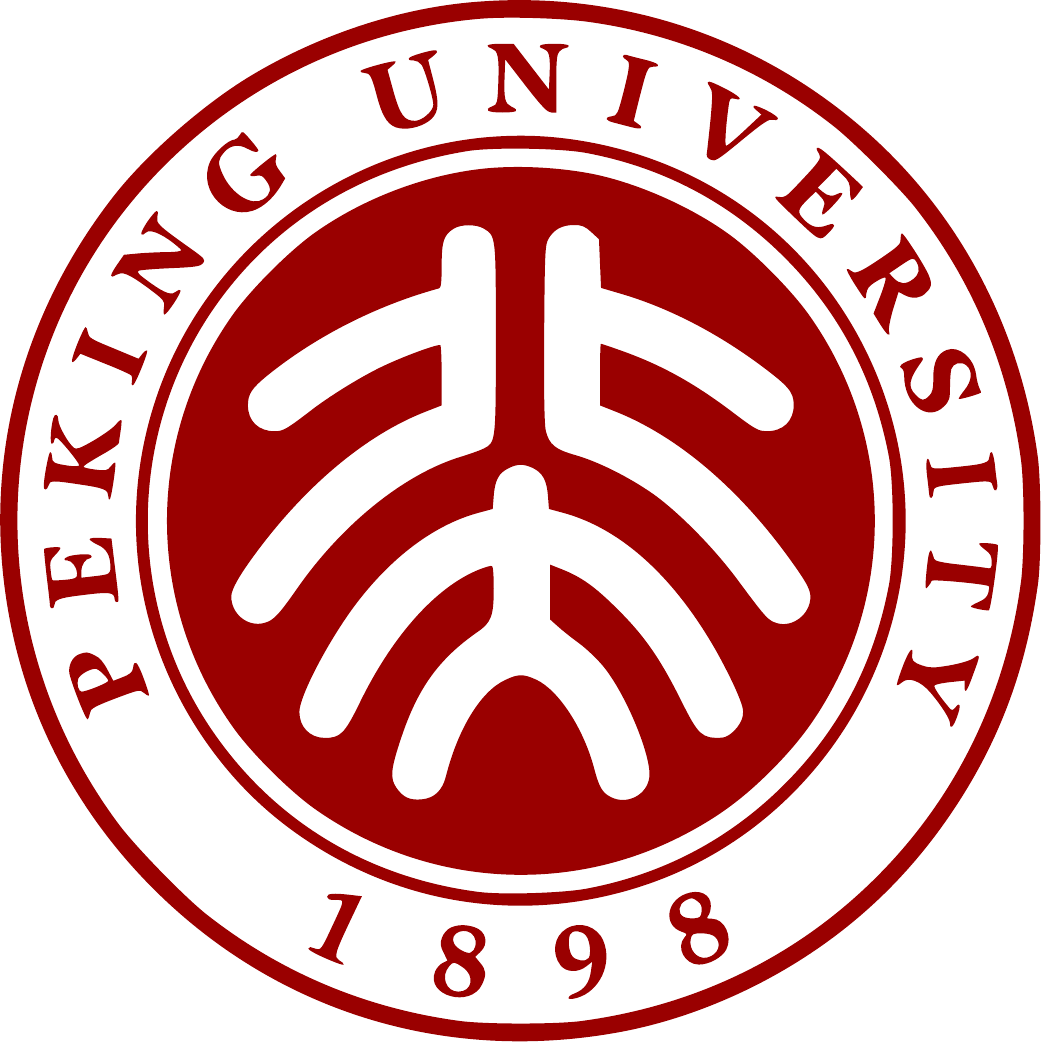}\hspace{2mm}%
                \includegraphics[height=10mm]{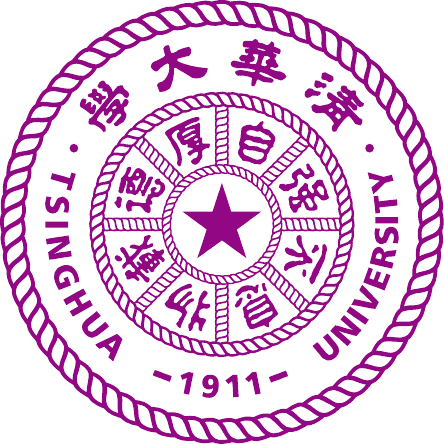}%
            }%
        }%
    }
}

\begin{document}
\maketitle

\vspace{-6mm}
\noindent
\begin{minipage}{\textwidth}
\centering
\includegraphics[width=0.96\textwidth]{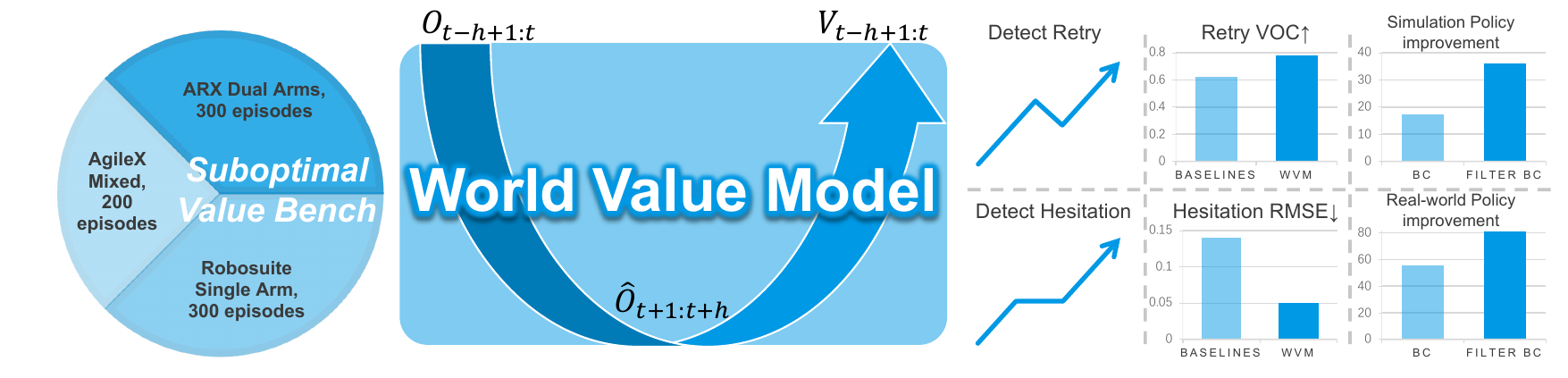}
\refstepcounter{figure}\label{fig:teaser}
\par\vspace{0.2em}
{\footnotesize\textbf{Figure \thefigure:} \textbf{Overview of World Value Models and Suboptimal-Value-Bench.}
\method leverages the world models to estimate temporally grounded task progress from videos, enabling the reliable detection of suboptimal behaviors such as hesitation and retry.
Evaluated on \bench, our new $800$-trajectory multi-embodiment benchmark, \method significantly enhances both high-fidelity value estimation and downstream policy learning.\par}
\end{minipage}
\vspace{0.3em}

\section{Introduction}
\label{sec:intro}

Generalist value models serve as a cornerstone for scaling robotic policy learning from vast and heterogeneous datasets~\cite{xvla,uniact,datamil,remix,mutual}, providing learning signals for large scale real-world RL systems~\cite{grrl,rl-token,sop} as well as offline data filters~\cite{pi06,pi07}. At its core, accurate value estimation requires a dual capability: a thorough comprehension of past temporal contexts~\cite{pomdp,recurrentQ,rpg,r2d2}, combined with prospective forward-looking planning over long-term future outcomes~\cite{alphago,dreamer,muzero,vpn}. However, merging these temporal dimensions into a single value estimator remains a formidable challenge in practice~\cite{longterm,lstmRL,creditsurvey,tvt,gtrxl}.

Existing value models, despite their notable promise, struggle to deliver high-fidelity progress estimations because they are hindered by three primary bottlenecks: (1) inefficiencies in value learning due to a heavy dependence on scalar value supervision~\cite{VLAC,robodopamine}, (2) limited generalization resulting from narrow, task-specific customization~\cite{grrl,rl-token,pi06,viva}, and (3) impaired temporal understanding and future planning capabilities, a direct consequence of the sparse visual modeling characteristic of underlying VLMs~\cite{VLAC,robodopamine,GVL,roboreward,topreward}.

World models have achieved notable success in modeling temporal dynamics and forecasting future states across both video generation~\cite{sora,seedance,wan22} and robotic manipulation~\cite{mimic-video,dreamzero,lingbot-va}, demonstrating strong capabilities in spatial and temporal understanding. Consequently, they inherently possess the dual properties required by a generalist value estimator. Driven by this, our key insight is that the spatiotemporal priors embedded in world models can be repurposed as a powerful foundation for value learning.
We introduce \method, a World Value Model engineered to inherit rich spatial-temporal priors from a pretrained video world model. Specifically, \method couples the video stream with a lightweight value Diffusion-Transformer (DiT)~\cite{dit} via a Mixture-of-Transformers (MoT)~\cite{mot} architecture. This design allows value tokens to selectively attend to structural video latents while minimizing representation interference with the video generation stream during training. To achieve comprehensive value learning over large-scale data corpus, \method formulates the value function as a distributional value chunk trained by flow matching~\cite{flow}. This formulation provides dense training signals and enhanced expressiveness, thereby yielding superior value estimation performance compared to traditional scalar supervision and conventional categorical distributions~\cite{hl-gaussian,c51}. Finally, a suite of augmentations, including video rewinding~\cite{rewind} and value prefix randomization, are applied during training, empowering \method to robust prediction both optimal and suboptimal task progress during inference.

When evaluating generalist value models, current practices mainly rely either on qualitative human visual comparisons or on non-scalable downstream policy performance. Another common choice is VOC~\cite{GVL}, but it can only reflect value models' task progress awareness on optimal trajectories. To address these limitations, we introduce \bench, a multi-embodiment benchmark comprising 800 trajectories paired with human-annotated ground-truth task progress. Featuring two prevalent suboptimal behavioral modes—retries and hesitations—\bench enables a comprehensive evaluation of generalist value models that extends far beyond the scope of existing metrics.

Experiments show that \method can generate values with higher qualities than value-model baselines on both \bench and expert VOC. Also the ablations validate the necessity of our world-model prior and core architectural choices. Beyond standalone value quality evaluation, integrating \method into downstream policy learning yields substantial performance gains with only noisy-data in both simulation and real-world manipulation tasks. We will make \method and \bench available to support the community.

Our main contributions are threefold:

(1) We repurpose world models as foundational backbones for robotic value learning, leveraging their rich spatiotemporal priors to overcome the limitations of standard VLMs.

(2) \method is the first large-scale value flow model that formulates value functions as distributional chunks. Complemented by simple-yet-effective design choices, \method achieves SOTA performance across diverse benchmarks and proven to be effective for policy improvement.

(3) We introduce \bench, a new evaluation suite featuring dense, human-labeled suboptimal trajectories tailored for value model evaluations.

\section{Related Work}
\label{sec:related}

\paragraph{Value models for robotic manipulation.}

Existing robotic value models face three persistent bottlenecks as generalist progress estimators.
First, scalar value regression on high-dimensional observations provides a sparse, low-information supervision signal, yielding sample-inefficient training and brittle predictions when scaled to heterogeneous video corpora~\cite{VLAC,robodopamine}.
Second, many value models~\cite{grrl,rl-token,pi06} are tightly tailored to a single task and thus cannot serve as a generalist progress estimator.
Third, generalist value estimators built on pretrained VLM backbones~\cite{VLAC,robodopamine,GVL,roboreward,topreward,robometer} inherit a representation optimized for static or temporally sparse images, and thus cannot capture dense temporal dynamics.
While ViVa~\cite{viva} represents the closest attempt to ours by building upon video models, it is confined to single-task settings and remains fundamentally restricted to action-annotated data.
To address these limitations, \method reformulates value estimation as a distributional chunk, enabling scalable, multi-task learning across massive, action-free video datasets by leveraging expressive visual-physical priors from a pretrained video world model.

\paragraph{World models for robotic manipulation.}

World models~\cite{world-models,world-models-survey,jepa} have recently gained prominent traction in robotic manipulation through the emergence of \emph{World Action Models} (WAMs)~\cite{dreamzero,wam-survey}, which jointly model action-conditioned visual dynamics. Prior works~\cite{mimic-video,dreamzero,lingbot-va,cosmos,motubrain,dit4dit,gigaworld} demonstrate that such video priors significantly boost learning efficiency by leveraging the world model's innate capacities for both temporal reasoning and forward prediction. Building upon this, Fast-WAM~\cite{fast-wam} illustrates that operating future predictions entirely within the latent space retains these representational benefits without the need for explicit pixel-level decoding~\cite{leworldmodel,lbp}. Shifting away from these predominantly policy-centric deployments, \method repurposes latent video priors for value estimation. Through a MoT design, our framework preserves the backbone's intrinsic video-modeling capacity while successfully deriving robust value predictions from its latent temporal features.

\paragraph{Evaluation of robotic value models.}

Quantifying the performance of robotic value models remains a critical challenge~\cite{GVL,robometer}. Early works~\cite{r3m,decisionnce,vip,liv} rely on qualitative curve inspection, which does not scale to systematic comparison. Another line of work~\cite{VLAC,robodopamine} measures value quality indirectly through downstream policy success, entangling value fidelity with policy choice and incurring substantial computational overhead. GVL~\cite{GVL} proposes \emph{Value-Order Correlation} (VOC), but its monotonicity criterion only applies to expert trajectories, and thus lacks assessment for suboptimal segments. Complementing all three, our \bench evaluates value models on human-annotated hesitation and retry trajectories, directly reflecting their ability to flag suboptimal segments.

\section{Method}
\label{sec:method}

\afterpage{%
\begin{figure*}[t]
\centering
\includegraphics[width=1.0\linewidth]{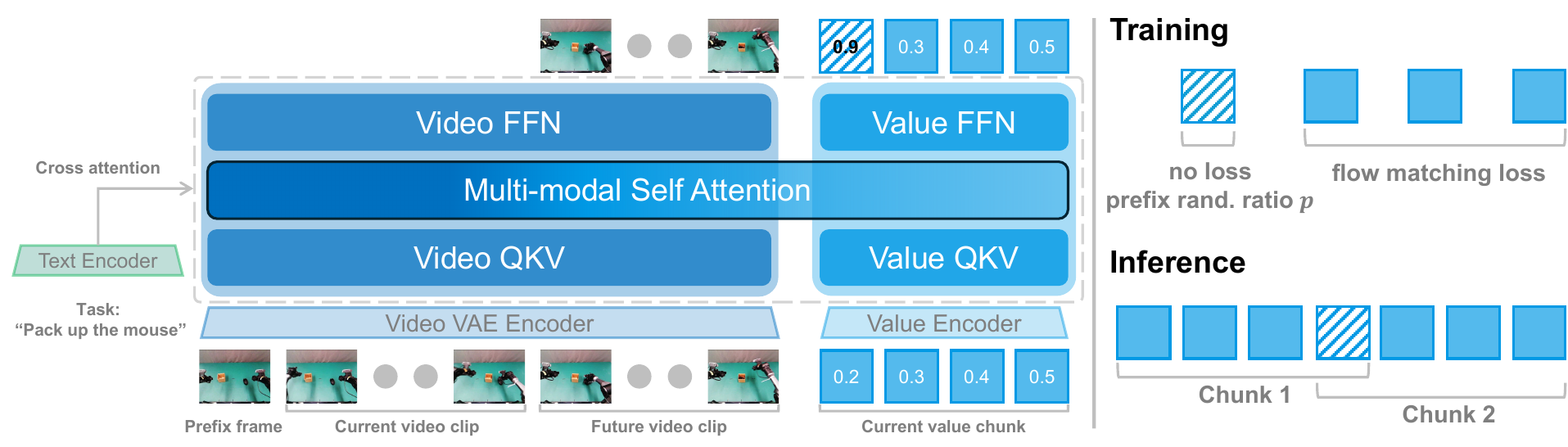}
\caption{\textbf{\method}'s architecture, prefix randomization and chunk overlapping scheme.}
\label{fig:arch}
\end{figure*}%
}

\subsection{Problem Formulation}

\label{sec:method_prelim}

We formulate value estimation as a chunk-wise prediction problem. Given an $h$-frame observation sequence $o_{t-h+1:t}$ and a language instruction $l$, the value model defines a conditional distribution over a length-$h$ sequence of per-frame values:
\begin{equation}
p_\psi(\hat{v}_{t-h+1:t} \mid o_{t-h+1:t},\, l),
\label{eq:value_model}
\end{equation}
where $\hat{v}_{t-h+1:t} \in [0,1]^h$. Here, $v_t = t / T$ denotes the normalized task progress, where $T$ is the total trajectory length. Modeling the full length-$h$ chunk instead of an isolated scalar enables the value model to capture local progress profiles—such as plateaus and regressions—that are pivotal for tracking temporal dynamics. Classically, a value function in reinforcement learning (RL) is defined as the expected discounted sum of future rewards~\cite{value,rl-intro}:
\begin{equation}
\label{eq:outcome}
V(o_t) = \mathbb{E}\left[\sum_{t'=t}^{T} \gamma^{t'-t} r_{t'} \;\middle|\; o_t\right],
\end{equation}
where $r_{t'}$ is the step-level reward and $\gamma \in (0,1]$ is the discount factor. Under the canonical sparse-reward setting where $r_{t'} = -1$ for non-terminal steps and $0$ at task completion, $V(o_t)$ reduces to the negative expected distance-to-goal, rendering value estimation equivalent to task-progress prediction. By construction, the value function thereby intrinsically focuses on future outcomes~\cite{interplay}. This perspective naturally motivates utilizing a video world model $M_\omega$ as a rich feature extractor for value estimation:
\begin{equation}
\label{eq:world_value_model}
p_\psi(\hat{v}_{t-h+1:t} \mid o_{t-h+1:t},\, l) = p_\psi(\hat{v}_{t-h+1:t} \mid M_\omega(o_{t-h+1:t},\, l)).
\end{equation}

\subsection{\method Architecture}

\label{sec:method_arch}

As illustrated in \figref{fig:arch}, \method instantiates \eqnref{eq:world_value_model} via a video DiT, a value DiT, and Multi-model self attention to enable the value stream capitalize temporal video features.

\paragraph{Video stream.}

\method uses the video VAE and video DiT of Wan2.2~\cite{wan22} as the world-modeling stream. For a value chunk anchored at the time window $[t-h+1,t]$, we first feed the video VAE a clean video clip with length $(2h+1)$ consisting of one prefix frame, the current observation frames, and the target future frames:
\begin{equation}
\underbrace{o_{t-h}}_{\text{1-frame prefix}} \;\Vert\; \underbrace{o_{t-h+1:t}}_{h\text{ current frames}} \;\Vert\; \underbrace{o_{t+1:t+h}}_{h\text{ future frames}}.
\end{equation}
The VAE encodes the clip into three temporal latents: we discard the prefix latent, keep the current latent as context, and corrupt the future latent for video-generation denoising.

\paragraph{Value stream and MoT coupling.}

The value stream is a lightweight DiT that mirrors the architecture of the video DiT with substantially fewer parameters. It predicts the value chunk $\hat{v}_{t-h+1:t}$ from noisy value tokens while attending to intermediate video-DiT features through MoT multi-modal self-attention. We adopt an asymmetric attention mask: value tokens attend to current video tokens but video tokens never attend to value tokens~\cite{fast-wam}.

\subsection{Training}

\label{sec:method_train}

\paragraph{Training objective.}

We apply flow matching~\cite{flow,rectifiedflow,flow-tutorial} to the supervised video and value tokens. Let $y$ denote either the future video latents $\xi_{t+1:t+h}$ or the value chunk $v_{t-h+1:t}$, and let $f_\psi$ denote the corresponding velocity predictor. We sample noise $\epsilon \sim \mathcal{N}(0, I)$ and a time step $\tau \in (0,1)$, construct the interpolated sample $y_\tau = \tau\, y + (1-\tau)\, \epsilon$, and train $f_\psi$ to predict the velocity field $y - \epsilon$:
\begin{equation}
\mathcal{L}_{\mathrm{FM}}(y) \;=\; \mathbb{E}_{y,\,\epsilon,\,\tau}\!\left[\,\big\| f_\psi(y_\tau,\,\tau,\,o_{t-h+1:t},\,l) - (y - \epsilon) \big\|_2^2 \,\right].
\label{eq:fm_loss}
\end{equation}
Instantiating \eqnref{eq:fm_loss} on the value chunk $v_{t-h+1:t}$ and the future video latents $\xi_{t+1:t+h}$ gives
\begin{equation}
\mathcal{L}_{\mathrm{value}} = \mathcal{L}_{\mathrm{FM}}(v_{t-h+1:t}),\qquad
\mathcal{L}_{\mathrm{video}} = \mathcal{L}_{\mathrm{FM}}(\xi_{t+1:t+h}).
\end{equation}
The overall objective is
\begin{equation}
\mathcal{L} \;=\; \mathcal{L}_{\mathrm{value}} + \lambda\, \mathcal{L}_{\mathrm{video}},
\end{equation}
where $\lambda$ controls the weight of video co-training; see \appref{app:impl} for details.

\paragraph{Prefix randomization.}

Inference with value chunk overlapping improves chunk continuity but can introduce a shortcut, allowing the value stream to extrapolate from this prefix without visual evidence. To prevent this behavior, we apply prefix randomization during training, analogous to the conditioning dropout in classifier-free guidance (CFG)~\cite{cfg,dipole}: with probability $p$, the prefix value is replaced by a random scalar sampled uniformly from $[0,1]$; otherwise, the prefix is retained. The loss is applied only to the remaining value tokens~\cite{training-time-rtc}. Mixing clean and randomized prefixes preserves inter-chunk continuity while preventing the value stream from taking the prefix as a shortcut. We ablate $p$ in \secref{sec:ablation_design}.

\paragraph{Video rewinding.}

Expert trajectories provide solely monotonic progress labels, offering limited supervision for plateaus or regressions. Following ReWiND~\cite{rewind}, we apply chunk-level rewind augmentation: for each window $o_{t-h+1:t}$, we sample one of three temporal patterns over the $h$ frames---\emph{rising}, \emph{plateau} or \emph{descending}---by preserving, repeating or reversing the frames, with $v_{t-h+1:t}$ relabeled accordingly. This exposes the value stream to local progress profiles associated with smooth advancement, hesitation and retry.

\begin{figure*}[t]
\centering
\includegraphics[width=\linewidth]{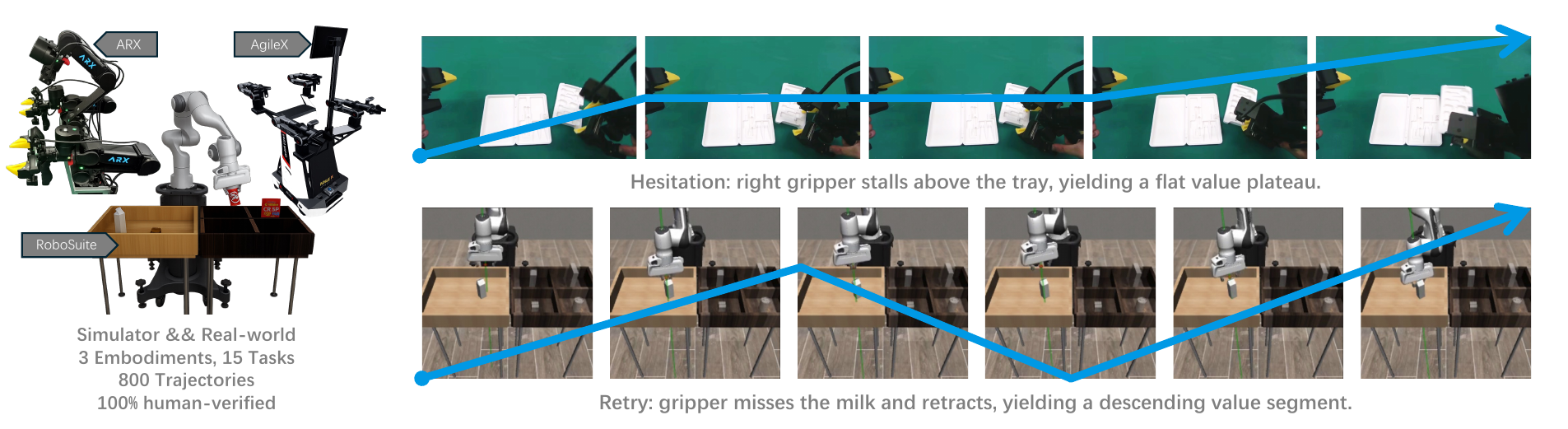}

\caption{Setup of \bench, and blue arrows (\textcolor{SkyBlue}{$\boldsymbol{\longrightarrow}$}) schematically illustrate the per-frame human-annotated value curve under suboptimal modes (hesitation and retry).}
\label{fig:bench}

\end{figure*}

\section{Suboptimal-Value-Bench}

\label{sec:bench}

Real robot datasets commonly exhibit suboptimal segments (e.g., hesitation and retry) that are pivotal for practical value estimation. We present \bench, a benchmark consisting of $800$ human-annotated trajectories across three embodiments and $15$ tasks, with each frame augmented by a dense value curve focusing on hesitation and retry (\autoref{fig:bench}). We refer the reader to \appref{app:bench_details} for comprehensive dataset details.

\subsection{Hesitation Segments}

\label{sec:bench_hesitation}

During a hesitation segment, the robot either remains stationary or executes task-irrelevant micro-movements rather than advancing the task progress. This behavior typically arises from teleoperator cognitive pauses (e.g., searching for a visual target) or physical hardware constraints (e.g., deceleration near joint limits). Therefore the task progress remains invariant throughout the segment. Because the standard VOC metric is ill-defined for invariant target trajectories~\cite{pearson}, we evaluate hesitation segments using the Root Mean Squared Error:
\begin{equation}
\text{Hesitation-RMSE} \;=\; \sqrt{\frac{1}{|\mathcal{H}|}\sum_{t \in \mathcal{H}} (\hat{v}_t - v_t)^2},
\end{equation}
where $\mathcal{H}$ denotes the set of frames within a hesitation segment, and $v_t$ represents the constant ground-truth value over that interval. This metric explicitly penalizes prediction drift; a model that maintains a constant, accurate prediction achieves an error of zero, whereas fluctuating predictions incur a higher RMSE proportional to their tracking deviation.

\subsection{Retry Segments}

\label{sec:bench_retry}

A retry episode is characterized by a failed manipulation attempt—such as an unsuccessful grasp—followed by a release and retraction phase prior to the subsequent attempt. Since capturing the resulting drop in value is paramount to identifying retries, our evaluation isolates temporal windows that exhibit monotonically decreasing ground-truth progress. We assess a value model's capacity to track this downward trend by restricting the VOC calculation exclusively to these windows, reporting the metric as Retry-VOC. A perfectly tracking, monotonically decreasing prediction yields a maximum score of $+1$, whereas an inverse, monotonically increasing prediction receives the worst-case score of $-1$.

\section{Experiments}

\label{sec:result}

We first analyze the quality of value predictions across both \bench and standard expert trajectories (\secref{sec:exp_value}). We then investigate whether \method can facilitate downstream policy acquisition (\secref{sec:exp_policy}). Finally, we conduct comprehensive ablation studies to dissect the contributions of our key design choices (\secref{sec:exp_ablation}).

\subsection{Value Estimation Quality}

\label{sec:exp_value}

\paragraph{Benchmarks and baselines.}

We evaluate value estimation performance on both \bench and standard VOC benchmarks. We compare our approach against six competitive baselines: GVL~\cite{GVL}, VLAC~\cite{VLAC}, Robometer~\cite{robometer}, TopReward~\cite{topreward}, RoboReward~\cite{roboreward}, and Robo-Dopamine~\cite{robodopamine}. Detailed standard VOC dataset settings and baseline implementation details are provided in \appref{app:expert_voc} and \appref{app:baselines}, respectively.

\begin{table*}[t]
    \centering
    \footnotesize
    \setlength{\tabcolsep}{5pt}
    \renewcommand{\arraystretch}{1.15}
    \resizebox{\textwidth}{!}{%
    \begin{tabular}{l cccccc c}
        \toprule
        \textbf{Hesitation ~ RMSE $\downarrow$}
        & \textbf{\textcolor{cGVL}{GVL}}
        & \textbf{\textcolor{cVLAC}{VLAC}}
        & \textbf{\textcolor{cRoboMeter}{Robometer}}
        & \textbf{\textcolor{cTopReward}{TopReward}}
        & \textbf{\textcolor{cRoboReward}{RoboReward}}
        & \textbf{\textcolor{cRoboDopamine}{Robo-Dopamine}}
        & \textbf{\method} \\
        \midrule
        Suboptimal-AgileX     & 0.11 & 0.47 & 0.13 & 0.36 & 0.12 & 0.41 & \textbf{0.07} \\
        Suboptimal-ARX        & 0.14 & 0.50 & 0.12 & 0.24 & 0.17 & 0.52 & \textbf{0.05} \\
        Suboptimal-RoboSuite  & 0.16 & 0.54 & 0.16 & 0.33 & 0.31 & 0.51 & \textbf{0.04} \\
        \midrule
        \textbf{Average}      & 0.14 & 0.51 & 0.14 & 0.31 & 0.21 & 0.49 & \textbf{0.05} \\
        \bottomrule
    \end{tabular}%
    }
    \caption{Evaluation of Hesitation-RMSE on \bench.}
    \label{tab:suboptimal_hesitation}
    
\end{table*}

\begin{savenotes}
\begin{table*}[b]
    \centering
    \scriptsize
    \setlength{\tabcolsep}{5pt}
    \renewcommand{\arraystretch}{1.15}
    \begin{tabular*}{\textwidth}{@{\extracolsep{\fill}} l cccc c}
        \toprule
        \textbf{Retry ~ VOC $\uparrow$}
        & \textbf{\textcolor{cGVL}{GVL}}
        & \textbf{\textcolor{cVLAC}{VLAC}}
        & \textbf{\textcolor{cRoboMeter}{Robometer}}
        & \textbf{\textcolor{cTopReward}{TopReward}}
        & \textbf{\method} \\
        \midrule
        Suboptimal-AgileX     & 0.73 & -0.37 &  0.32 &  0.15 & \textbf{0.79} \\
        Suboptimal-ARX        & 0.76 & /\footnote{``/'' marks an ill-defined VOC, which holds consistently for RoboReward and Robo-Dopamine.} & -0.27 & -0.19 & \textbf{0.79} \\
        Suboptimal-RoboSuite  & 0.43 & / & -0.37 &  0.00 & \textbf{0.75} \\
        \midrule
        \textbf{Average}      & 0.62 & -0.37 & -0.16 &  0.00 & \textbf{0.78} \\
        \bottomrule
    \end{tabular*}
    \caption{Evaluation of Retry-VOC on \bench.}
    \label{tab:suboptimal_retry}
\end{table*}
\end{savenotes}

\paragraph{Performance on Suboptimal-Value-Bench.}

\begin{figure*}[t]
\centering
\includegraphics[width=\linewidth]{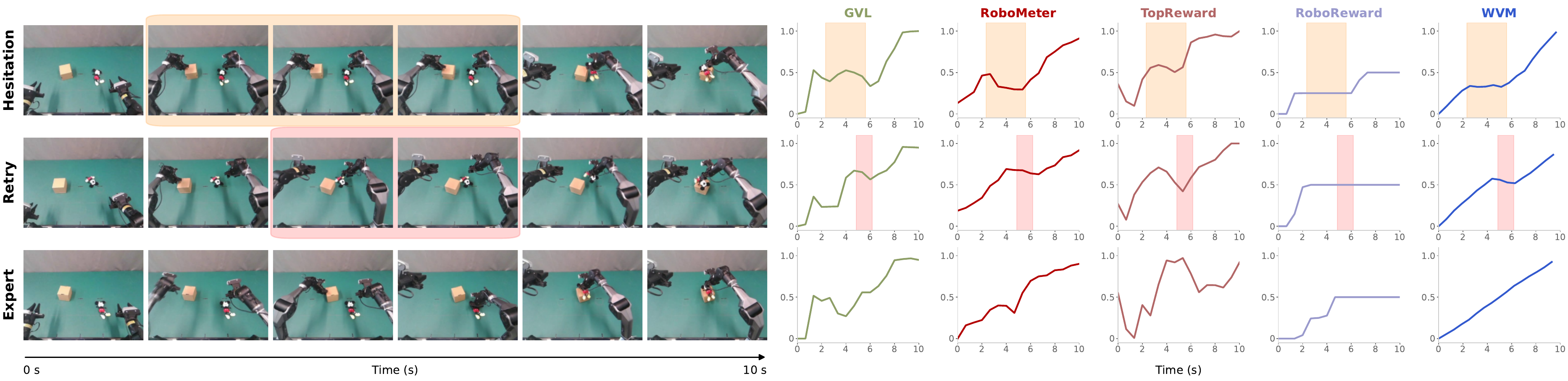}
\caption{\textbf{Qualitative value comparison results.} From top to bottom: (a) Hesitation: gripper pauses before catching the doll; (b) Retry: gripper backtracks after an unsuccessful grasp; (c) Expert: gripper successfully catches the doll and puts it on the box.}
\label{fig:eye_comparison}

\end{figure*}

As shown in \autoref{tab:suboptimal_hesitation}, \method consistently achieves the lowest Hesitation-RMSE across all three embodiments. Crucially, it reduces the average error to $0.05$, outperforming the strongest baselines, GVL and Robometer (both $0.14$), by a substantial margin. These findings validate that \method provides superior value estimation stability, effectively mitigating prediction drift during periods of task-invariant stagnation.
Similarly, \autoref{tab:suboptimal_retry} reveals a comparable performance advantage for \method within retry phases. Specifically, our method secures the top Retry-VOC score across all embodiments, boosting the average metric from $0.62$ to $0.78$. Finally, qualitative comparisons in \autoref{fig:eye_comparison} show that the hesitation and retry segments flagged by \method align closely with human intuition, consistent with its superior performance on \bench.

\paragraph{Performance on Expert-VOC.}

As shown in \autoref{tab:expert_voc}, \method also leads on clean expert trajectories, achieving the highest average VOC score of $0.95$ versus $0.88$ for the strongest baseline and ranking first on five of the six datasets. It exceeds $0.99$ on all three self-collected datasets, confirming strong monotonic tracking on clean demonstrations. The sole exception is EgoDex, where RoboReward slightly outperforms \method ($0.95$ vs.\ $0.92$), exposing limitations of Expert-VOC as a metric for value models that we revisit in \secref{sec:ablation_design}.

\begin{table*}[t]
    \centering
    \footnotesize
    \setlength{\tabcolsep}{5pt}
    \renewcommand{\arraystretch}{1.15}
    \resizebox{\textwidth}{!}{%
    \begin{tabular}{l cccccc c}
        \toprule
        \textbf{Expert ~ VOC $\uparrow$}
        & \textbf{\textcolor{cGVL}{GVL}}
        & \textbf{\textcolor{cVLAC}{VLAC}}
        & \textbf{\textcolor{cRoboMeter}{Robometer}}
        & \textbf{\textcolor{cTopReward}{TopReward}}
        & \textbf{\textcolor{cRoboReward}{RoboReward}}
        & \textbf{\textcolor{cRoboDopamine}{Robo-Dopamine}}
        & \textbf{\method} \\
        \midrule
        OXE                         & 0.67 & 0.48 & 0.63 & 0.19 & 0.92 & 0.72 & \textbf{0.94} \\
        RoboCOIN                    & 0.70 & 0.60 & 0.77 & 0.47 & 0.85 & 0.75 & \textbf{0.95} \\
        EgoDex                      & 0.82 & 0.62 & 0.86 & 0.37 & \textbf{0.95} & 0.88 & 0.92 \\
        Self-collected (3 embodiments) & 0.93 & 0.50 & 0.93 & 0.58 & 0.84 & 0.76 & \textbf{0.99} \\
        \midrule
        \textbf{Average}            & 0.78 & 0.59 & 0.81 & 0.42 & 0.88 & 0.82 & \textbf{0.95} \\
        \bottomrule
    \end{tabular}%
    }
    \caption{Value-Order Correlation on expert demonstrations.}
    \label{tab:expert_voc}
    
\end{table*}

\subsection{Downstream Policy Learning}

\label{sec:exp_policy}

\paragraph{Experimental setup.}

\begin{figure*}[b]
\centering
\includegraphics[width=\linewidth]{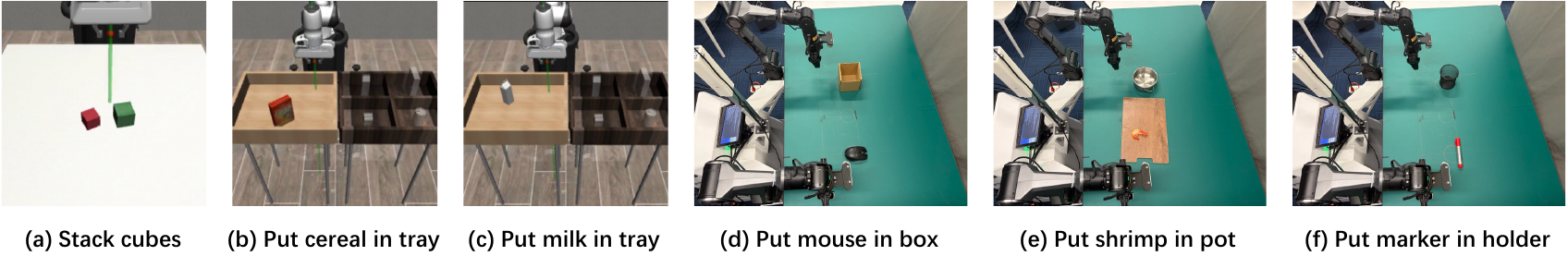}
\caption{Task setups for downstream policy learning experiments.}
\label{fig:setup}

\end{figure*}

We evaluate downstream policy learning across three simulated RoboSuite tasks and three real-world AgileX bimanual manipulation tasks, with their experimental setups illustrated in \autoref{fig:setup}. We employ $\pi_{0.5}$-base~\cite{pi05} as our foundational policy. For policy finetuning, we exclusively utilize \textit{suboptimal} data, consisting of only 10 trajectories for each simulated task and 50 trajectories for each real-world task. Additional implementation details are provided in \appref{app:policy_details}.

\paragraph{Policy improvement.}

\begin{figure*}[t]
\centering
\includegraphics[width=\textwidth]{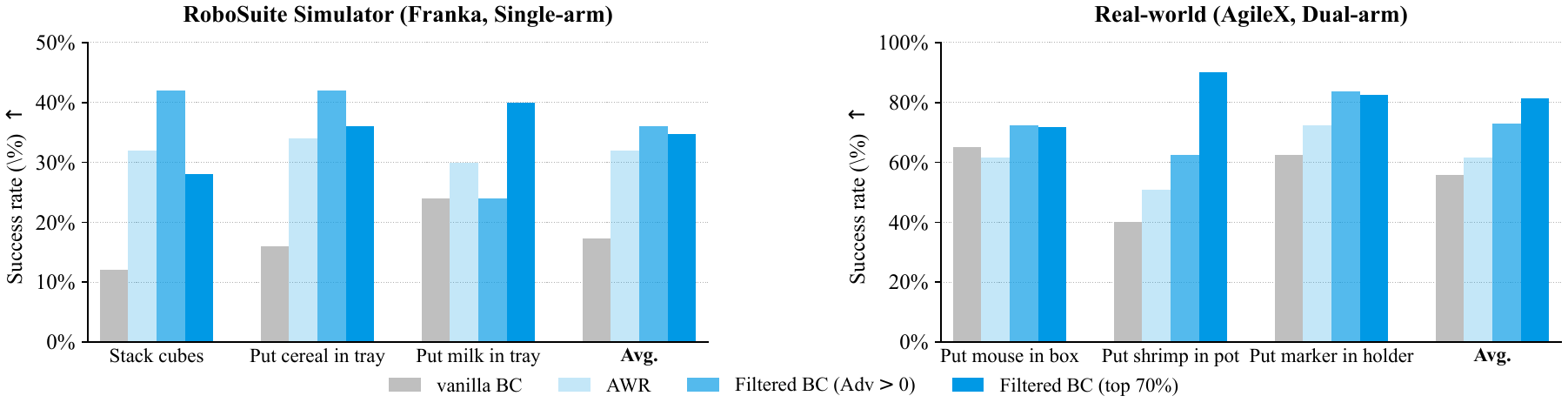}
\caption{Policy improvement results by combining \method with AWR and Filtered BC.}
\label{fig:policy_results}

\end{figure*}

Building upon vanilla Behavioral Cloning (BC), we evaluate Advantage Weighted Regression (AWR)~\cite{awr,awr-control} alongside two variants of Filtered BC: a binary filter that exclusively retains trajectory segments with positive advantages~\cite{grrl}, and a percentile filter that preserves the top 70\% of segments ranked by \method values~\cite{pi06}. As demonstrated in \autoref{fig:policy_results}, all three \method-guided variants consistently outperform the vanilla BC baseline across both simulated and physical environments. These experimental results validate that \method can truly distinguish genuine task progress from suboptimal behaviors, thereby enabling the downstream policy to leverage imperfect data more effectively.

\subsection{Ablation Study}

\label{sec:exp_ablation}
\begin{table*}[t]
    \centering
    \footnotesize
    \setlength{\tabcolsep}{6pt}
    \renewcommand{\arraystretch}{1.15}
    \resizebox{\textwidth}{!}{%
    \begin{tabular}{l c ccc cc c}
        \toprule
        \multirow{2}{*}[-0.4\baselineskip]{\textbf{Metric}}
        & \multirow{2}{*}[-0.4\baselineskip]{\textbf{Ours}}
        & \multicolumn{3}{c}{Video DiT variants}
        & \multicolumn{2}{c}{Prefix rand.}
        & Value head \\
        \cmidrule(lr){3-5} \cmidrule(lr){6-7} \cmidrule(lr){8-8}
        & & w/o $\mathcal{L}_{\mathrm{video}}$ & scratch & frozen & $p{=}0$ & $p{=}1$ & HL-Gaussian \\
        \midrule
        Hesitation-RMSE $\downarrow$  & \textbf{0.05} & 0.08 & 0.08 & 0.12 & 0.09 & \textbf{0.05} & 0.06 \\
        Retry-VOC $\uparrow$         & \textbf{0.78} & 0.68 & 0.62 & 0.45 & 0.67 & 0.75 & 0.59 \\
        Expert-VOC $\uparrow$        & 0.95 & 0.95 & 0.96 & 0.92 & \textbf{0.98} & 0.91 & 0.87 \\
        \bottomrule
    \end{tabular}%
    }
    \caption{Ablation results of \method's design choices.}
    \label{tab:ablation}
    
\end{table*}
To evaluate key design choices of \method, we conduct ablation studies and summarize results in \autoref{tab:ablation}. Unless specified otherwise, other components maintain the default configurations.

\paragraph{Video co-training ($\lambda$).}
\label{sec:ablation_video}

Omitting the video co-training objective ($\mathcal{L}_{\mathrm{video}}$) consistently degrades performance across all \bench metrics. Specifically, Hesitation-RMSE increases from $0.05$ to $0.08$ and Retry-VOC drops from $0.78$ to $0.68$, even though the pretrained video DiT continues to be optimized via the value-gradient pathway. Furthermore, training the video stream from random initialization reduces Retry-VOC to $0.62$, whereas completely freezing the video weights yields the most severe performance degradation ($0.12$ Hesitation-RMSE and $0.45$ Retry-VOC). These empirical findings underscore that explicit video co-training is indispensable for the value model to capture underlying temporal dynamics, thereby validating the central premise of \method: a continuously co-trained video world model serves as a principled and effective backbone for value estimation.

\paragraph{Prefix randomization ($p$).}
\label{sec:ablation_design}

The prefix randomization ratio $p$ manages the trade-off between temporal shortcut suppression and inter-chunk continuity. Without randomization ($p{=}0$), Hesitation-RMSE worsens to $0.09$ and Retry-VOC drops to $0.67$, whereas Expert-VOC increases to $0.98$. This divergence exposes a pathological over-reliance on prefixes as a causal shortcut, confirming Expert-VOC is an insufficient standalone metric for value models. Conversely, full masking ($p{=}1$) recovers retry detection ($0.75$) but drops Expert-VOC to $0.91$ due to disrupted cross-chunk consistency. \method's $p{=}0.5$ achieves the optimal balance, maximizing robustness on \bench and expert trajectories.

\paragraph{Value head design.}
\label{sec:ablation_head}

Distributional value heads are widely adopted in value learning~\cite{grrl,pi06,hl-gaussian,c51}. Replacing our flow-matching head with an HL-Gaussian alternative (\appref{app:hl-gaussian}) degrades all metrics, with a marginal increase in Hesitation-RMSE ($0.05 \to 0.06$) and a sharp decline in discriminative scores. This underscores a core limitation of categorical heads: their fixed, pre-specified bin support preserves the conditional mean but discards the fine-grained density variations needed by ordinal metrics. Conversely, our flow-matching head captures a continuous return density without bounding the support or resolution, retaining the local value differentials critical for ranking temporally adjacent chunks~\cite{value_flows}.

\section{Conclusion}

\label{sec:conclusion}

We introduce \method, a generalist robotic value flow model rooted in the predictive capabilities of a pretrained world model. Inheriting its native strengths in historical grounding and future planning, \method delivers SOTA results across both standard benchmarks and our new \bench—a multi-embodiment suite of $800$ high-fidelity suboptimal trajectories with frame-level human annotations, complementing expert-only evaluations. Downstream deployments in simulation and reality confirm that this world-model-derived architecture offers robust and effective guidance for learning from mixed-quality data.

\section{Limitations}

\label{sec:limitations}

While \method demonstrates strong performance, several limitations remain. First, due to computational constraints, our training dataset is currently limited in scale; consequently, \method exhibits restricted zero-shot capacity when confronted with entirely unseen tasks and scenes. Second, although \bench broadens evaluation beyond expert-only demonstrations, its scope is primarily focused on pick-and-place tasks. Expanding the benchmark to more dexterous~\cite{xvla,grrl} and long-horizon manipulations~\cite{mem,physiagent} is a critical next step. We plan to scale up both training mixture and evaluation diversity in future work.

\clearpage
\beginappendix
\appendix

\renewcommand{\thesection}{\Alph{section}}
\renewcommand{\thefigure}{\thesection.\arabic{figure}}
\renewcommand{\thetable}{\thesection.\arabic{table}}
\renewcommand{\theequation}{\thesection.\arabic{equation}}
\setcounter{figure}{0}
\setcounter{table}{0}
\setcounter{equation}{0}

\section{Implementation Details}
\label{app:impl}

\paragraph{Architecture.}

We build the video stream of \method upon the publicly released \textsc{Wan2.2-TI2V-5B} checkpoint~\cite{wan22}.
The accompanying Wan2.2-VAE compresses input videos by a factor of $4 \times 16 \times 16$ along the temporal and spatial axes, yielding a $48$-channel spatiotemporal latent. This latent is further patchified with a patch size of $(1,2,2)$ before entering the transformer.
The video DiT is a $30$-layer Diffusion Transformer with a hidden dimension of $3072$ ($24$ attention heads, head dimension $128$) and an FFN width of $14336$, totaling approximately $5.0$B parameters.
The value DiT shares the same depth but operates at a reduced hidden width of $512$, configured with $8$ self/cross-attention heads (head dimension $64$) and a matching FFN width of $14336$.
At each layer, the two streams are coupled via Mixture-of-Transformers (MoT) self-attention: video tokens retain the inherited Wan2.2 query/key/value projections, whereas value tokens are linearly projected from $512$ to the shared attention width of $3072$ ($24$ heads, head dimension $128$) to participate in joint attention, before being projected back to $512$ at the layer output.
Crucially, the MoT attention mask is asymmetric, enabling value tokens to attend to the video latents while ensuring the video stream remains completely unaffected by the value stream.
The additional value-side components (projections, cross-attention, FFN, adaptive-norm tables, and input/output heads) contribute roughly $0.7$B trainable parameters, bringing the full two-stream transformer to approximately $5.7$B parameters.
During training, both streams are optimized jointly, while the Wan2.2-VAE is frozen as a tokenizer and the T5 text encoder is executed offline to precompute static text embeddings.

\paragraph{Training.}

The complete set of hyperparameters used for the main \method run is summarised in \autoref{tab:training_hparams}.

\begin{table*}[h]
\centering
\footnotesize
\setlength{\tabcolsep}{6pt}
\renewcommand{\arraystretch}{1.1}
\begin{tabular}{l l}
\toprule
\textbf{Category} & \textbf{Setting} \\
\midrule
Hardware                        & $32\times$ NVIDIA A100-SXM4-$40$\,GB \\
Wall-clock training time        & $\sim 40$ hours \\
Optimizer                       & AdamW \\
$\beta_1$                       & $0.9$ \\
$\beta_2$                       & $0.95$ \\
Weight decay                    & $0$ \\
Gradient clipping (max norm)    & $1.0$ \\
Peak learning rate              & $1\!\times\!10^{-4}$ \\
LR schedule                     & Cosine decay to $0.1\times$ peak \\
Warm-up steps                   & $500$ \\
Global batch size               & $1024$ \\
Total training steps            & $30{,}000$ \\
Mixed precision                 & bfloat16 (bf16) \\
\midrule
Prefix randomization ratio $p$  & $0.5$ \\
Rewind ratio                    & $0.5$ \\
Rewind plateau ratio            & $0.1$ \\
Value chunk length $h$          & $4$ \\
Latent target FPS               & $2.0$ ($3.0$ for AgileX / ARX self-collected data) \\
Video co-training weight $\lambda$ & $1.0$ \\
\bottomrule
\end{tabular}
\caption{Training hyperparameters used for the main \method run.}
\label{tab:training_hparams}

\end{table*}

\paragraph{Inference.}

At test time, \method denoises the value chunk with an explicit Euler solver applied to the learned flow-matching velocity field, and we use only a single denoising step for all reported results. Empirically, using more denoising steps does not yield measurable gains on either \bench or Expert-VOC. We attribute this to the regime in which \method operates: relative to its model capacity, the training corpus is moderate in size, so the learned velocity field is sufficiently smooth that one Euler step already lands close to the ground-truth value chunk, and additional refinement provides no further signal. Inference is performed with a chunk size of $h{=}4$ and the overlapping-window averaging scheme described in \secref{sec:method_arch}.

\paragraph{Training dataset mixture.}

The composition of the training mixture used to pre-train \method is summarised in \autoref{tab:training_data}. We report each source along two granularities: ``Subsets'' counts the distinct subsets enumerated within the source, while ``Trajectories'' counts demonstrations. The semantics of a subset depends on the source: for the four self-collected sources (RoboSuite, AgileX single-arm, AgileX dual-arm, ARX) , one subset corresponds to exactly one task; for RoboCOIN, EgoDex, and RoboReward, one subset may bundle one or several tasks---specifically, a scene--language pair for RoboCOIN, a top-level task category covering many distinct language instructions for EgoDex, and a constituent sub-dataset for RoboReward. The same convention applies to \autoref{tab:expert_voc_data}.

\begin{savenotes}
\begin{table*}[h]
\centering
\footnotesize
\renewcommand{\arraystretch}{1.15}
\begin{tabular*}{\linewidth}{@{\extracolsep{\fill}} l c r r r c}
\toprule
\textbf{Data source} & \textbf{Type} & \textbf{Subsets} & \textbf{Trajectories} & \textbf{Hours} & \textbf{FPS} \\
\midrule
RoboCOIN~\cite{robocoin}\footnote{We use the snapshot of all RoboCOIN data available as of 2026-01-12.}     & Real-world & 432 &  98{,}171 &  673.80 & 30, 50 \\
EgoDex~\cite{egodex}         & Real-world & 111 & 299{,}100 &  688.56 & 30     \\
RoboReward~\cite{roboreward}\footnote{We use the train split of the RoboReward open-source release and retain only the successful demonstrations with the maximum reward label of $5$. For the DROID portion, which provides both left- and right-camera views, we keep only the left view.} & Real-world &  29 &   7{,}428 &   36.01 & 10     \\
RoboSuite (ours)             & Simulation &   6 &   1{,}865 &   11.32 & 10     \\
AgileX single-arm (ours)     & Real-world &   4 &      160  &    0.39 & 15     \\
AgileX dual-arm (ours)       & Real-world &   3 &      120  &    0.26 & 15     \\
ARX (ours)                   & Real-world &   5 &      242  &    0.50 & 15     \\
\midrule
\textbf{Total}               & ---        & \textbf{590} & \textbf{407{,}086} & \textbf{1{,}410.83} & --- \\
\bottomrule
\end{tabular*}
\caption{Composition of the training mixture used to pre-train \method. ``Subsets'' uses the source-specific semantics defined in the surrounding text. ``FPS'' lists the native frame rates available in each source before resampling.}
\label{tab:training_data}

\end{table*}
\end{savenotes}

\section{Suboptimal-Value-Bench Details}
\label{app:bench_details}

\paragraph{Per-task statistics.}

\bench covers three embodiments (AgileX, ARX, RoboSuite) and $15$ manipulation tasks, with each task instantiated as two trajectory groups corresponding to the two suboptimal modes (hesitation and retry). The full per-task breakdown is provided in \autoref{tab:bench_per_task}.

\begin{table*}[t]
\centering
\footnotesize
\renewcommand{\arraystretch}{1.15}
\setlength{\tabcolsep}{5pt}
\resizebox{\linewidth}{!}{%
\begin{tabular}{l l c c c c c}
\toprule
\textbf{Embodiment} & \textbf{Task} & \textbf{Arm} & \textbf{Hesitation} & \textbf{Retry} & \textbf{Total} & \textbf{Duration (min)} \\
\midrule
\multirow{5}{*}{AgileX (real)}
    & carrot off plate                    & single &  25 &  25 &  50 &  9.9 \\
    & carrot on plate                     & single &  25 &  25 &  50 & 10.1 \\
    & mickey box                          & dual   &  25 &  25 &  50 & 11.1 \\
    & sausage pot                         & dual   &  25 &  25 &  50 & 11.0 \\
    \cmidrule(lr){2-7}
    & \textit{Subtotal}                   & ---    & \textit{100} & \textit{100} & \textit{200} & \textit{42.0} \\
\midrule
\multirow{6}{*}{ARX (real)}
    & flip bottles                        & dual   &  15 &  15 &  30 &  5.1 \\
    & open box                            & dual   &  20 &  20 &  40 &  7.4 \\
    & split cups                          & dual   &  15 &  15 &  30 &  5.9 \\
    & stack bowls                         & dual   &  50 &  50 & 100 & 16.3 \\
    & stack cups                          & dual   &  50 &  50 & 100 & 16.6 \\
    \cmidrule(lr){2-7}
    & \textit{Subtotal}                   & ---    & \textit{150} & \textit{150} & \textit{300} & \textit{51.3} \\
\midrule
\multirow{7}{*}{RoboSuite (sim)}
    & lift                                & single &  25 &  25 &  50 & 14.0 \\
    & pick \& place bread                 & single &  25 &  25 &  50 & 20.4 \\
    & pick \& place can                   & single &  25 &  25 &  50 & 23.3 \\
    & pick \& place cereal                & single &  25 &  25 &  50 & 23.3 \\
    & pick \& place milk                  & single &  25 &  25 &  50 & 21.3 \\
    & stack                               & single &  25 &  25 &  50 & 17.6 \\
    \cmidrule(lr){2-7}
    & \textit{Subtotal}                   & ---    & \textit{150} & \textit{150} & \textit{300} & \textit{119.9} \\
\midrule
\textbf{Total} & --- & --- & \textbf{400} & \textbf{400} & \textbf{800} & \textbf{213.2} \\
\bottomrule
\end{tabular}%
}
\caption{Per-task composition of \bench. Each task contributes two trajectory groups, one per suboptimal mode (hesitation and retry); the columns ``Hesitation'' and ``Retry'' report the trajectory count in each group.}
\label{tab:bench_per_task}

\end{table*}

\paragraph{Annotation tool.}

Manually labelling every frame of $800$ trajectories is prohibitively expensive. To accelerate annotation while preserving label quality, we adopt a two-stage pipeline. In the first stage, each trajectory is fed to a proprietary large vision-language model through its public API to obtain an initial coarse segmentation of non-progress intervals. Frames are sampled at a downsampled frame rate, prefixed with their index, and submitted together with the natural-language task description using the following prompt:

\begin{quote}
\footnotesize
\begin{verbatim}
Below are frames sampled from a {fps:.0f}fps robot manipulation video
(total {frames[-1][0]} frames). Each frame is preceded by its frame
index label [frame=N].

{frames_block}

The task is: "{task_description}".

Please analyze the full trajectory and identify segments where the
robot is NOT making forward progress toward completing the task.
A segment is "not forward progress" if during that time the task
state is stalled or regressing, for example:
- The robot is stuck, hesitating, or repeating a motion without
  advancing the task.
- The task state is going backwards (e.g. an object was dropped,
  knocked away, or released unnecessarily).
- The robot is performing motions that do not bring it closer to
  task completion.

Do NOT flag segments where the robot is actively and successfully
advancing toward the goal, even if the motion is slow.

Output a JSON object with the following format (NO extra text
outside the JSON):
{
  "non_progress_segments": [
    {
      "start_frame": <int>,
      "end_frame": <int>,
      "description": "<brief description of why this segment is
                      not forward progress>"
    }
  ],
  "task_completed": <true or false>,
  "summary": "<one-sentence summary>"
}

If the entire trajectory is efficient with no wasted time, return
an empty non_progress_segments list.
\end{verbatim}
\end{quote}

\begin{figure}[h]
\centering
\includegraphics[width=0.95\linewidth]{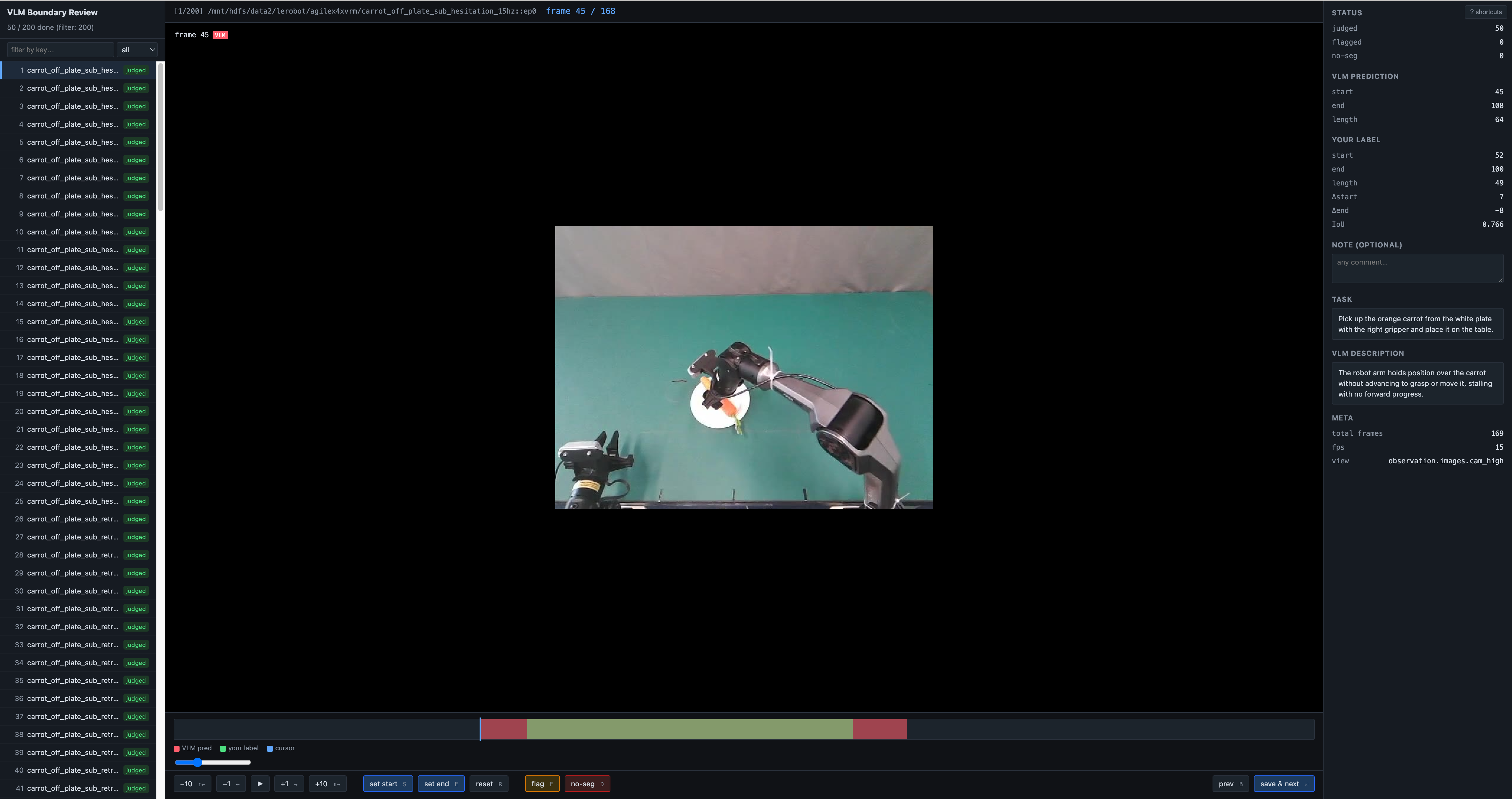}
\caption{Human verification interface used to refine the VLM-proposed non-progress intervals. Annotators can replay the trajectory video, adjust segment boundaries at frame-level resolution.}
\label{fig:human_verifier}

\end{figure}

\noindent The model returns, for every trajectory, a set of candidate non-progress intervals together with a one-sentence rationale. In the second stage, human annotators inspect each candidate interval in a custom labelling interface (\autoref{fig:human_verifier}), correct its boundaries at frame-level resolution. The interface displays the trajectory video together with a frame-aligned timeline of the VLM-proposed segments, and allows annotators to drag segment endpoints, split or merge intervals, and re-play any sub-clip. Trajectories whose VLM proposal is empty are still presented to annotators in full to avoid silent omissions. This pre-segmentation step substantially reduces the search effort of human labellers while leaving all final boundaries and labels under human control.

\paragraph{Per-frame label to ground-truth value.}

For every trajectory the human annotation provides a suboptimal type (\textit{hesitation} or \textit{retry}) together with three integers: the inclusive endpoints $m,n$ ($0<m<n<T-1$) of the suboptimal segment and the total trajectory length $T$. We then construct the per-frame ground-truth value $v_t \in [0,1]$ as a four-point piecewise-linear curve through the control points
\begin{equation}
(0,\,0),\quad (m,\,v_m),\quad (n,\,v_n),\quad (T-1,\,1),
\end{equation}
with $v_t$ obtained by linear interpolation between adjacent control points (frames outside $[0,T-1]$ are clamped to $0$ or $1$). Let $x = n-m$ denote the length of the suboptimal segment. The values $v_m,v_n$ are chosen so that the resulting curve reflects the semantics of each suboptimal mode:

\textit{Hesitation.} During $[m,n]$ the robot does not advance the task---it either remains stationary or executes task-irrelevant micro-movements---and then resumes at the same effective speed, so the segment is a plateau in progress space and the remaining $T-1-x$ effective frames evenly cover the unit progress:
\begin{equation}
v_m \;=\; v_n \;=\; \frac{m}{T-1-x}.
\end{equation}
The three slopes become $\tfrac{1}{T-1-x}$, $0$, and $\tfrac{1}{T-1-x}$, giving a piecewise curve that advances at a constant effective rate before and after the plateau.

\textit{Retry.} We model the retry as a uniform-speed retraction: during $[m,n]$ the robot moves backward at the same per-frame rate $r$ at which it normally advances forward, an assumption we find well supported by the retry trajectories in our data. Out of the $T$ trajectory frames, $x$ are spent retracting and a further $x$ forward frames are needed to re-earn the lost progress, leaving only $T-2x$ frames of net forward motion to cover the unit interval; hence $r = 1/(T-2x)$. Concretely,
\begin{equation}
v_m \;=\; \frac{m}{T-2x},
\qquad
v_n \;=\; \max\!\Bigl(0,\;\tfrac{m-x}{T-2x}\Bigr).
\end{equation}
When the retry happens late enough that $n \le 2m$, the three segments share the same absolute slope $\tfrac{1}{T-2x}$ with the middle segment negative, producing a symmetric ``V'' over the retry interval. When the retry is unusually long ($n>2m$), $v_n$ would otherwise be negative and is clamped to $0$, so the curve drops linearly from $v_m$ to $0$ and then re-climbs to $1$ over the remaining $(T-1-n)$ frames; this preserves the semantics that progress cannot fall below zero.

We use these piecewise-linear curves as the ground truth for both the Hesitation-RMSE and Retry-VOC metrics in \secref{sec:bench}.

\section{Expert VOC Dataset Composition}
\label{app:expert_voc}

The composition of the evaluation set used for Expert-VOC is summarised in \autoref{tab:expert_voc_data}. It covers three public corpora (OXE, RoboCOIN, EgoDex) and four self-collected expert sets that match the embodiments used in \bench. All trajectories are held out from the \method training mixture in \autoref{tab:training_data} at the trajectory level to avoid leakage.

\begin{table*}[h]
\centering
\footnotesize
\renewcommand{\arraystretch}{1.15}
\begin{tabular*}{\linewidth}{@{\extracolsep{\fill}} l c r r r c}
\toprule
\textbf{Data source} & \textbf{Type} & \textbf{Subsets} & \textbf{Trajectories} & \textbf{Hours} & \textbf{FPS} \\
\midrule
RoboCOIN~\cite{robocoin}        & Real-world & 318 &  641 & 3.02 & 30 \\
EgoDex~\cite{egodex}            & Real-world &  13 &  588 & 1.04 & 30 \\
OXE~\cite{oxe}                  & Real-world &  10 &  146 & 0.58 & 10 \\
AgileX single-arm (ours)        & Real-world &   4 &   40 & 0.10 & 15 \\
AgileX dual-arm (ours)          & Real-world &   3 &   30 & 0.07 & 15 \\
ARX (ours)                      & Real-world &   5 &   61 & 0.13 & 15 \\
RoboSuite (ours)                & Simulation &   6 &   99 & 0.60 & 10 \\
\midrule
\textbf{Total} & --- & \textbf{359} & \textbf{1{,}605} & \textbf{5.52} & --- \\
\bottomrule
\end{tabular*}
\caption{Composition of the evaluation set used for Expert-VOC. ``Subsets'' follows the same source-specific semantics as in \autoref{tab:training_data} (see the ``Training dataset mixture'' paragraph for the per-source definition). The four self-collected sources cover the same three embodiments (AgileX, ARX, RoboSuite) used by \bench, and are reported as a single ``Self-collected (3 embodiments)'' row in the main paper (\autoref{tab:expert_voc}).}
\label{tab:expert_voc_data}

\end{table*}

\paragraph{Overall protocol.}

Unless stated otherwise, the Expert-VOC pool for each data source is obtained by uniformly sampling $5\%$ of its trajectories at random; the remaining $95\%$ are retained in the \method training mixture (\autoref{tab:training_data}), so the two splits are disjoint at the trajectory level and the held-out set is never seen during training. The source-specific deviations from this default are detailed below.

\paragraph{OXE~\cite{oxe}.}

Our OXE trajectories are drawn from the RoboReward~\cite{roboreward} open-source release rather than the raw OXE distribution; concretely, we re-use exactly the same filtered pool (train split, reward $=5$ successes, DROID left view only) described in the footnote of \autoref{tab:training_data}, and the per-source selection protocol below operates on top of that pool. OXE itself aggregates many sub-datasets whose trajectory quality varies substantially, so a uniform random $5\%$ sample over this pool would still pull low-quality trajectories into the expert set. We adopt the per-dataset VOC ranking reported by GVL~\cite{GVL} as a quality proxy. GVL observes that several large sub-datasets, most notably DROID~\cite{droid}, are ranked very low, consistent with prior reports that removing DROID from large action-model training improves final performance; on inspection of the low-VOC DROID clips, GVL attributes the low scores to poor camera placement that fails to capture the robot's motion and to heavy occlusions of the arm and the manipulated object, both of which would directly bias an expert-progress evaluator. We therefore restrict OXE to the ten sub-datasets with the highest VOC scores reported by GVL, and then apply the default $5\%$ random sample within those sub-datasets, yielding the $146$ expert trajectories listed in \autoref{tab:expert_voc_data}.

\paragraph{AgileX and ARX (ours).}

The self-collected AgileX (single-arm and dual-arm) and ARX trajectories are teleoperated by trained operators and are uniformly of high quality, so no additional quality filtering is needed. To obtain a sufficiently large Expert-VOC pool for stable per-embodiment estimates given the small absolute size of these sources, we raise the sampling ratio from the default $5\%$ to $20\%$ for all three subsets.

\paragraph{RoboSuite (ours).}

RoboSuite trajectories are generated by a scripted oracle policy and are uniformly expert, so no quality filtering is required either. Because RoboSuite already contributes a large number of trajectories, the default $5\%$ random sample alone produces a sufficiently large expert pool ($99$ trajectories), and we apply no additional adjustment.

\paragraph{RoboCOIN~\cite{robocoin}.}

RoboCOIN provides multiple successful demonstrations per language instruction and the per-demonstration execution speed varies substantially. Following the common heuristic that, among successful demonstrations of the same instruction, the faster ones tend to be closer to expert behaviour, we enumerate the distinct language instructions in RoboCOIN and, for each instruction, keep the single shortest-duration trajectory as the Expert-VOC sample (rather than using the default random $5\%$).

\paragraph{EgoDex~\cite{egodex}.}

EgoDex covers a wide range of egocentric human activities, many of which are unrelated to robotic manipulation. We therefore first shortlist the EgoDex task categories that correspond to basic pick-and-place manipulation, since these are the activities most directly relevant to manipulation-oriented value estimation. Within each shortlisted category we then enumerate the distinct language instructions and, for each instruction, keep the single shortest demonstration, using the same speed-as-expertise heuristic as for RoboCOIN.

\section{Value Model Baseline Reproduction}
\label{app:baselines}

All six baselines are evaluated under a unified video-sampling pipeline: each trajectory is first downsampled to a common target frame rate ($2$\,fps, or $3$\,fps for the AgileX and ARX datasets to keep per-trajectory frame counts comparable across embodiments), and we then evaluate each baseline using its officially recommended sampling protocol (multi-anchor prefix evaluation for \emph{Robometer} and \emph{RoboReward}; single-pass full-trajectory evaluation for the remaining four). Our evaluation harness directly builds on the public Robometer~\cite{robometer} codebase.

\paragraph{GVL~\cite{GVL}.}

GVL casts value estimation as autoregressive completion-percentage prediction over \emph{shuffled} video frames, where a close-source VLM is prompted with a task description together with a sequence of frame--index pairs and asked to output a task-progress percentage for each frame. We use the public API of \texttt{gpt-5.4} as the backbone, with a per-call frame budget of $32$ frames.

\paragraph{VLAC~\cite{VLAC}.}

VLAC fine-tunes an InternVL backbone into a unified action--critic model that, given a pair of image observations and a language goal, autoregressively emits a signed progress delta together with a task-completion (done) signal, providing dense rewards for downstream RL. We use the public checkpoint \texttt{InternRobotics/VLAC} in single-pass mode with $32$ frames per trajectory and the default decoding configuration (temperature $0.5$, batch size $32$); we keep the original \texttt{frame\_skip} schedule and disable the auxiliary image branch (\texttt{use\_images=false}).

\paragraph{Robometer~\cite{robometer}.}

Robometer fine-tunes Qwen3-VL-4B with a composite objective combining a frame-level progress loss, a per-frame success loss, and an inter-trajectory preference loss, enabling it to learn from both expert and suboptimal/failed trajectories and to emit dense per-frame progress estimates over short prefix clips. We use the public checkpoint \texttt{robometer/Robometer-4B} with the official multi-anchor evaluation: $5$ uniformly spaced anchors per trajectory, $8$ frames per anchor (\texttt{max\_frames=8}, \texttt{use\_frame\_steps=true}, \texttt{subsample\_n\_frames=5}), and a model batch size of $32$.

\paragraph{RoboReward~\cite{roboreward}.}

RoboReward fine-tunes Qwen3-VL at the 4B and 8B scales to predict a discrete end-of-episode progress score in $\{1,\dots,5\}$ from a task description and a full rollout video; following the original setup, dense per-frame rewards are obtained by re-querying the model on partial-trajectory prefixes. We use the public checkpoint \texttt{teetone/RoboReward-4B} with the same multi-anchor protocol as Robometer ($5$ anchors, \texttt{use\_frame\_steps=true}, \texttt{subsample\_n\_frames=5}) but raise \texttt{max\_frames} to $32$ and cap the generation length at $128$ tokens, matching the original setup.

\paragraph{TopReward~\cite{topreward}.}

TopReward turns a frozen video VLM into a zero-shot temporal value function by reading off the log-probability that the model assigns to an affirmative completion token (e.g., \texttt{True}) when asked whether a video prefix has completed the instruction; in our setup, we follow the official protocol and mean-aggregate this score over $K$ prefix samples per trajectory. We use \texttt{Qwen/Qwen3-VL-8B-Instruct} as the backbone with $K=15$ prefix samples, mean reduction, the official chat template, and a $2$\,fps temporal sampling; each trajectory is consumed once with \texttt{max\_frames=32}.

\paragraph{Robo-Dopamine~\cite{robodopamine}.}

Robo-Dopamine is a 3B-parameter step-aware generative process reward model (the GRM in the Robo-Dopamine framework) that, conditioned on a task description and multi-view ``before''/``after'' frames, predicts a discretised relative progress hop and supports an incremental inference mode that emits a per-frame progress signal as frames are streamed in. We use the public checkpoint \texttt{tanhuajie2001/Robo-Dopamine-GRM-3B} in incremental mode (\texttt{eval\_mode="incremental"}, \texttt{frame\_interval=1}, batch size $1$) and feed up to $32$ uniformly sampled frames per trajectory in a single pass.

\section{Downstream Policy Learning Details}
\label{app:policy_details}

The hyperparameters used for all downstream policy-learning experiments in \secref{sec:exp_policy} are summarised in \autoref{tab:policy_hparams}.

\paragraph{Chunk-level advantage proxy.}

All weighting schemes operate on the same chunk-level advantage proxy derived from \method.  For a sample anchored at frame $t$ with action-chunk length $H$, let $t^{\mathrm{head}}=t$ and $t^{\mathrm{tail}}=\min(t+H-1,\,T-1)$, and let $V(\cdot)\in[0,1]$ denote the per-frame value emitted by \method.  We define
\begin{equation}
\Delta_i \;=\; V(t^{\mathrm{tail}}_i) \;-\; V(t^{\mathrm{head}}_i),
\end{equation}
which approximates the value improvement contributed by the action chunk and plays the role of an advantage estimate in the weighting schemes below.  Given per-sample BC losses $\ell_i$ (e.g.\ flow-matching loss for $\pi_{0.5}$-base) and weights $w_i$, the weighted-BC objective used throughout this section is
\begin{equation}
\mathcal{L} \;=\; \frac{\sum_{i=1}^{B} w_i\, \ell_i}{\sum_{i=1}^{B} w_i + \varepsilon},
\qquad \varepsilon = 10^{-6}.
\end{equation}

\paragraph{Filtered BC.}

The two Filtered-BC variants in \secref{sec:exp_policy} both use a hard-threshold indicator on $\Delta_i$:
\begin{equation}
w_i \;=\; \mathbf{1}\!\left[\,\Delta_i \;\ge\; \kappa\,\right].
\end{equation}
The \textit{binary} variant uses $\kappa=0.0$, which simply discards chunks on which the value does not improve.  The \textit{percentile} variant retains the top $70\%$ of chunks ranked by $\Delta_i$ over the training set; the corresponding threshold $\kappa$ is reported per task suite in \autoref{tab:policy_hparams}.

\paragraph{Advantage-Weighted Regression (AWR).}

AWR replaces the hard indicator with a clipped exponential weight on the same advantage proxy:
\begin{equation}
w_i \;=\; \min\!\Bigl(\exp\!\bigl(\tau\cdot \Delta_i\bigr),\; \delta\Bigr).
\end{equation}
This recovers the standard AWR/AWAC weighting~\cite{awr,awr-control} with temperature $1/\beta\!=\!\tau$, advantage $A\!=\!\Delta_i$, and clipping ceiling $M\!=\!\delta$.  We fix $\delta=2.0$ across both task suites and adjust the temperature $\tau$ per suite (see \autoref{tab:policy_hparams}) to match the empirical scale of $\Delta_i$ at each suite's action-chunk length $H$.  The clip caps the highest-advantage chunks at $2{\times}$ their baseline contribution, preventing a few high-advantage outliers from dominating the gradient.  We further renormalise the per-batch weights so that $\frac{1}{B}\sum_i w_i = 1$, keeping the gradient scale aligned with vanilla BC and allowing us to reuse the same learning rate and weight decay as the BC baseline.

\begin{table*}[h]
\centering
\footnotesize
\setlength{\tabcolsep}{6pt}
\renewcommand{\arraystretch}{1.1}
\begin{tabular}{l l}
\toprule
\textbf{Category} & \textbf{Setting} \\
\midrule
\multicolumn{2}{l}{\textit{Shared}} \\
\midrule
Base policy                     & $\pi_{0.5}$-base \\
Hardware                        & $16\times$ NVIDIA A100-SXM4-$40$\,GB \\
Optimizer                       & AdamW \\
$\beta_1$                       & $0.9$ \\
$\beta_2$                       & $0.95$ \\
Peak learning rate              & $2.5\!\times\!10^{-5}$ \\
Global batch size               & $256$ \\
AWR clip ceiling $\delta$       & $2.0$ \\
Filtered-BC (binary) threshold $\kappa$ & $0.0$ \\
\midrule
\multicolumn{2}{l}{\textit{Task-suite specific (RoboSuite / AgileX)}} \\
\midrule
Control mode                                & EEF / Joint \\
Total training steps                        & $5{,}000$ / $10{,}000$ \\
Wall-clock training time                    & $\sim 4.5$\,h / $\sim 9$\,h \\
Action-chunk length $H$                     & $10$ / $50$ \\
AWR temperature $\tau$                      & $10$ / $2$ \\
Filtered-BC (top-$70\%$) threshold $\kappa$ & $0.02$ / $0.06$ \\
\bottomrule
\end{tabular}
\caption{Hyperparameters used for downstream policy fine-tuning in \secref{sec:exp_policy}. The action-chunk length, AWR temperature, and top-$70\%$ Filtered-BC threshold are set per task suite to account for the different per-frame value-change scales induced by different chunk lengths.}
\label{tab:policy_hparams}

\end{table*}

For evaluation, each RoboSuite task is rolled out for $50$ trials, while each real-world AgileX task is rolled out for $30$ trials, and we report the average success rate over these trials.

\section{HL-Gaussian Value Head Details}
\label{app:hl-gaussian}

The HL-Gaussian ablation in \secref{sec:ablation_head} replaces \method's flow-matching value head with a discrete distributional head~\cite{hl-gaussian} that reformulates per-frame value regression as classification over $K$ fixed bins. All non-head components (video stream, value stream, MoT coupling, prefix randomization, video rewinding) are kept identical to \method, so the only varying factor is the value head itself.

\paragraph{Bin support and soft targets.}

We discretise the support $[0,1]$ with $K=51$ equally-spaced bin centres $c_k = (k-1)/(K-1)$ for $k=1,\dots,K$. Given a ground-truth value $v\in[0,1]$, the soft target distribution is a Gaussian-smoothed one-hot over the bins:
\begin{equation}
p_k(v) \;=\; \frac{\exp\!\left(-(v - c_k)^2 / 2\sigma^2\right)}{\sum_{j=1}^{K} \exp\!\left(-(v - c_j)^2 / 2\sigma^2\right)},
\qquad
\sigma \;=\; \frac{1}{K-1}.
\end{equation}
We set $\sigma$ to one bin width by default, which gives smooth-but-locally-peaked targets; values within $10^{-6}$ of the endpoints $0$ or $1$ are snapped to a hard one-hot at bin $1$ or bin $K$ to avoid leaking probability mass onto neighbouring bins at the boundaries.

\paragraph{Training objective.}

The value DiT outputs $K$ logits $z\in\mathbb{R}^K$ at every (frame, sub-step) token. We train it with a token-level soft cross-entropy against the encoded target distribution, which is equivalent to minimising the KL divergence up to a constant entropy term:
\begin{equation}
\mathcal{L}_{\mathrm{value}}^{\mathrm{HLG}} \;=\; \mathbb{E}_{v}\!\left[\,-\sum_{k=1}^{K} p_k(v)\,\log\,\mathrm{softmax}(z)_k\,\right].
\end{equation}
This replaces the flow-matching loss $\mathcal{L}_{\mathrm{value}}$ in \eqnref{eq:fm_loss}; the video co-training loss $\mathcal{L}_{\mathrm{video}}$ and the overall objective weight $\lambda$ are kept unchanged from \method.

\paragraph{Inference.}

Unlike the flow-matching head, the HL-Gaussian head does \emph{not} require any iterative denoising: a single transformer forward yields $K$ logits per (frame, sub-step) token, and we decode the predicted scalar as the expectation of the softmax distribution over the bin centres:
\begin{equation}
\hat{v} \;=\; \sum_{k=1}^{K} \mathrm{softmax}(z)_k \cdot c_k.
\end{equation}
We then apply the same overlapping-window averaging scheme described in \secref{sec:method_arch} to assemble per-frame value predictions across adjacent chunks.

\clearpage
\bibliographystyle{plainnat}
\bibliography{ref}

@inproceedings{GVL,
  title={Vision language models are in-context value learners},
  author={Ma, Yecheng Jason and Hejna, Joey and Fu, Chuyuan and Shah, Dhruv and Liang, Jacky and Xu, Zhuo and Kirmani, Sean and Xu, Peng and Driess, Danny and Xiao, Ted and others},
  booktitle={The Thirteenth International Conference on Learning Representations},
  year={2024}
}

@article{robometer,
  title     = {Robometer: Scaling General-Purpose Robotic Reward Models via Trajectory Comparisons},
  author={Anthony Liang and Yigit Korkmaz and Jiahui Zhang and Minyoung Hwang and Abrar Anwar and Sidhant Kaushik and Aditya Shah and Alex S. Huang and Luke Zettlemoyer and Dieter Fox and Yu Xiang and Anqi Li and Andreea Bobu and Abhishek Gupta and Stephen Tu and Erdem Biyik and Jesse Zhang},
  year={2026},
  journal   = {arXiv preprint arXiv:2603.02115}
}

@article{roboreward,
  title={RoboReward: General-Purpose Vision-Language Reward Models for Robotics},
  author={Lee, Tony and Wagenmaker, Andrew and Pertsch, Karl and Liang, Percy and Levine, Sergey and Finn, Chelsea},
  journal={arXiv preprint arXiv:2601.00675},
  year={2026}
}

@article{topreward,
  title={Topreward: Token probabilities as hidden zero-shot rewards for robotics},
  author={Chen, Shirui and Harrison, Cole and Lee, Ying-Chun and Yang, Angela Jin and Ren, Zhongzheng and Ratliff, Lillian J and Duan, Jiafei and Fox, Dieter and Krishna, Ranjay},
  journal={arXiv preprint arXiv:2602.19313},
  year={2026}
}

@article{VLAC,
  title={A vision-language-action-critic model for robotic real-world reinforcement learning},
  author={Zhai, Shaopeng and Zhang, Qi and Zhang, Tianyi and Huang, Fuxian and Zhang, Haoran and Zhou, Ming and Zhang, Shengzhe and Liu, Litao and Lin, Sixu and Pang, Jiangmiao},
  journal={arXiv preprint arXiv:2509.15937},
  year={2025}
}

@article{viva,
  title={ViVa: A Video-Generative Value Model for Robot Reinforcement Learning},
  author={Lv, Jindi and Li, Hao and Li, Jie and Nie, Yifei and Kong, Fankun and Wang, Yang and Wang, Xiaofeng and Zhu, Zheng and Ni, Chaojun and Deng, Qiuping and others},
  journal={arXiv preprint arXiv:2604.08168},
  year={2026}
}

@article{robodopamine,
  title={Robo-Dopamine: General Process Reward Modeling for High-Precision Robotic Manipulation},
  author={Tan, Huajie and Chen, Sixiang and Xu, Yijie and Wang, Zixiao and Ji, Yuheng and Chi, Cheng and Lyu, Yaoxu and Zhao, Zhongxia and Chen, Xiansheng and Co, Peterson and others},
  journal={arXiv preprint arXiv:2512.23703},
  year={2025}
}

@article{xvla,
  title={X-vla: Soft-prompted transformer as scalable cross-embodiment vision-language-action model},
  author={Zheng, Jinliang and Li, Jianxiong and Wang, Zhihao and Liu, Dongxiu and Kang, Xirui and Feng, Yuchun and Zheng, Yinan and Zou, Jiayin and Chen, Yilun and Zeng, Jia and others},
  journal={arXiv preprint arXiv:2510.10274},
  year={2025}
}

@article{pi06,
  title={{$\pi^{*}_{0.6}$}: a VLA That Learns From Experience},
  author={Intelligence, Physical and Amin, Ali and Aniceto, Raichelle and Balakrishna, Ashwin and Black, Kevin and Conley, Ken and Connors, Grace and Darpinian, James and Dhabalia, Karan and DiCarlo, Jared and others},
  journal={arXiv preprint arXiv:2511.14759},
  year={2025}
}

@article{pi07,
  title={{$\pi^{*}_{0.7}$}: a Steerable Generalist Robotic Foundation Model with Emergent Capabilities},
  author={Intelligence, Physical and Ai, Bo and Amin, Ali and Aniceto, Raichelle and Balakrishna, Ashwin and Balke, Greg and Black, Kevin and Bokinsky, George and Cao, Shihao and Charbonnier, Thomas and others},
  journal={arXiv preprint arXiv:2604.15483},
  year={2026}
}

@article{grrl,
  title={Gr-rl: Going dexterous and precise for long-horizon robotic manipulation},
  author={Li, Yunfei and Ma, Xiao and Xu, Jiafeng and Cui, Yu and Cui, Zhongren and Han, Zhigang and Huang, Liqun and Kong, Tao and Liu, Yuxiao and Niu, Hao and others},
  journal={arXiv preprint arXiv:2512.01801},
  year={2025}
}

@article{cosmos,
  title={Cosmos Policy: Fine-Tuning Video Models for Visuomotor Control and Planning},
  author={Kim, Moo Jin and Gao, Yihuai and Lin, Tsung-Yi and Lin, Yen-Chen and Ge, Yunhao and Lam, Grace and Liang, Percy and Song, Shuran and Liu, Ming-Yu and Finn, Chelsea and Gu, Jinwei},
  journal={arXiv preprint arXiv:2601.16163},
  year={2026}
}

@misc{wam-survey,
      title={World Action Models: The Next Frontier in Embodied AI}, 
      author={Siyin Wang and Junhao Shi and Zhaoyang Fu and Xinzhe He and Feihong Liu and Chenchen Yang and Yikang Zhou and Zhaoye Fei and Jingjing Gong and Jinlan Fu and Mike Zheng Shou and Xuanjing Huang and Xipeng Qiu and Yu-Gang Jiang},
      year={2026},
      eprint={2605.12090},
      archivePrefix={arXiv},
      primaryClass={cs.RO},
      url={https://arxiv.org/abs/2605.12090}, 
}

@article{mot,
  title={Mixture-of-Transformers: A Sparse and Scalable Architecture for Multi-Modal Foundation Models},
  author={Weixin Liang and LILI YU and Liang Luo and Srini Iyer and Ning Dong and Chunting Zhou and Gargi Ghosh and Mike Lewis and Wen-tau Yih and Luke Zettlemoyer and Xi Victoria Lin},
  journal={Transactions on Machine Learning Research},
  issn={2835-8856},
  year={2025},
  url={https://openreview.net/forum?id=Nu6N69i8SB},
  note={}
}

@article{lingbot-va,
  title={Causal World Modeling for Robot Control},
  author={Li, Lin and Zhang, Qihang and Luo, Yiming and Yang, Shuai and Wang, Ruilin and Han, Fei and Yu, Mingrui and Gao, Zelin and Xue, Nan and Zhu, Xing and Shen, Yujun and Xu, Yinghao},
  journal={arXiv preprint arXiv:2601.21998},
  year={2026}
}

@article{mimic-video,
  author    = {Jonas Pai and Liam Achenbach and Victoriano Montesinos and Benedek Forrai and Oier Mees and Elvis Nava},
  title     = {mimic-video: Video-Action Models for Generalizable Robot Control Beyond VLAs},
  journal   = {arXiv preprint 2512.15692},
  year      = {2025},
}

@misc{dreamzero,
      title={World Action Models are Zero-shot Policies}, 
      author={Seonghyeon Ye and Yunhao Ge and Kaiyuan Zheng and Shenyuan Gao and Sihyun Yu and George Kurian and Suneel Indupuru and You Liang Tan and Chuning Zhu and Jiannan Xiang and Ayaan Malik and Kyungmin Lee and William Liang and Nadun Ranawaka and Jiasheng Gu and Yinzhen Xu and Guanzhi Wang and Fengyuan Hu and Avnish Narayan and Johan Bjorck and Jing Wang and Gwanghyun Kim and Dantong Niu and Ruijie Zheng and Yuqi Xie and Jimmy Wu and Qi Wang and Ryan Julian and Danfei Xu and Yilun Du and Yevgen Chebotar and Scott Reed and Jan Kautz and Yuke Zhu and Linxi "Jim" Fan and Joel Jang},
      year={2026},
      eprint={2602.15922},
      archivePrefix={arXiv},
      primaryClass={cs.RO},
      url={https://arxiv.org/abs/2602.15922}, 
}

@article{fast-wam,
  title={Fast-WAM: Do World Action Models Need Test-time Future Imagination?},
  author={Tianyuan Yuan and Zibin Dong and Yicheng Liu and Hang Zhao},
  journal={arXiv preprint arXiv:2603.16666},
  year={2026},
  url={https://arxiv.org/abs/2603.16666}
}

@inproceedings{decisionnce,
  title={DecisionNCE: Embodied Multimodal Representations via Implicit Preference Learning},
  author={Li, Jianxiong and Zheng, Jinliang and Zheng, Yinan and Mao, Liyuan and Hu, Xiao and Cheng, Sijie and Niu, Haoyi and Liu, Jihao and Liu, Yu and Liu, Jingjing and others},
  booktitle={Forty-first International Conference on Machine Learning},
  year={2024}
}

@article{r3m,
  title={R3m: A universal visual representation for robot manipulation},
  author={Nair, Suraj and Rajeswaran, Aravind and Kumar, Vikash and Finn, Chelsea and Gupta, Abhinav},
  journal={arXiv preprint arXiv:2203.12601},
  year={2022}
}

@article{wan22,
  title={Wan: Open and advanced large-scale video generative models},
  author={Wan, Team and Wang, Ang and Ai, Baole and Wen, Bin and Mao, Chaojie and Xie, Chen-Wei and Chen, Di and Yu, Feiwu and Zhao, Haiming and Yang, Jianxiao and others},
  journal={arXiv preprint arXiv:2503.20314},
  year={2025}
}

@article{training-time-rtc,
  title={Training-time action conditioning for efficient real-time chunking},
  author={Black, Kevin and Ren, Allen Z and Equi, Michael and Levine, Sergey},
  journal={arXiv preprint arXiv:2512.05964},
  year={2025}
}

@inproceedings{rewind,
    title={ReWi{ND}: Language-Guided Rewards Teach Robot Policies without New Demonstrations},
    author={Jiahui Zhang and Yusen Luo and Abrar Anwar and Sumedh Anand Sontakke and Joseph J Lim and Jesse Thomason and Erdem Biyik and Jesse Zhang},
    booktitle={9th Annual Conference on Robot Learning},
    year={2025},
    url={https://openreview.net/forum?id=XjjXLxfPou}
}

@article{pi05,
  title={{$\pi_{0.5}$}: a Vision-Language-Action Model with Open-World Generalization},
  author={Intelligence, Physical and Black, Kevin and Brown, Noah and Darpinian, James and Dhabalia, Karan and Driess, Danny and Esmail, Adnan and Equi, Michael and Finn, Chelsea and Fusai, Niccolo and others},
  journal={arXiv preprint arXiv:2504.16054},
  year={2025}
}

@article{hl-gaussian,
  title={Stop regressing: Training value functions via classification for scalable deep rl},
  author={Farebrother, Jesse and Orbay, Jordi and Vuong, Quan and Ta{\"\i}ga, Adrien Ali and Chebotar, Yevgen and Xiao, Ted and Irpan, Alex and Levine, Sergey and Castro, Pablo Samuel and Faust, Aleksandra and others},
  journal={arXiv preprint arXiv:2403.03950},
  year={2024}
}

@article{value_flows,
  title={Value flows},
  author={Dong, Perry and Zheng, Chongyi and Finn, Chelsea and Sadigh, Dorsa and Eysenbach, Benjamin},
  journal={arXiv preprint arXiv:2510.07650},
  year={2025}
}

@article{dipole,
  title={Dichotomous Diffusion Policy Optimization},
  author={Ruiming Liang and Yinan Zheng and Kexin Zheng and Tianyi Tan and Jianxiong Li and Liyuan Mao and Zhihao Wang and Guang Chen and Hangjun Ye and Jingjing Liu and Jinqiao Wang and Xianyuan Zhan},
  journal={arXiv preprint arXiv:2601.00898},
  year={2026}
}

@inproceedings{oxe,
  title={Open x-embodiment: Robotic learning datasets and rt-x models: Open x-embodiment collaboration 0},
  author={O’Neill, Abby and Rehman, Abdul and Maddukuri, Abhiram and Gupta, Abhishek and Padalkar, Abhishek and Lee, Abraham and Pooley, Acorn and Gupta, Agrim and Mandlekar, Ajay and Jain, Ajinkya and others},
  booktitle={2024 IEEE International Conference on Robotics and Automation (ICRA)},
  pages={6892--6903},
  year={2024},
  organization={IEEE}
}

@article{robocoin,
  title={RoboCOIN: An Open-Sourced Bimanual Robotic Data COllection for INtegrated Manipulation},
  author={Wu, Shihan and Liu, Xuecheng and Xie, Shaoxuan and Wang, Pengwei and Li, Xinghang and Yang, Bowen and Li, Zhe and Zhu, Kai and Wu, Hongyu and Liu, Yiheng and others},
  journal={arXiv preprint arXiv:2511.17441},
  year={2025}
}

@article{egodex,
  title={Egodex: Learning dexterous manipulation from large-scale egocentric video},
  author={Hoque, Ryan and Huang, Peide and Yoon, David J and Sivapurapu, Mouli and Zhang, Jian},
  journal={arXiv preprint arXiv:2505.11709},
  year={2025}
}

@article{vip,
  title={Vip: Towards universal visual reward and representation via value-implicit pre-training},
  author={Ma, Yecheng Jason and Sodhani, Shagun and Jayaraman, Dinesh and Bastani, Osbert and Kumar, Vikash and Zhang, Amy},
  journal={arXiv preprint arXiv:2210.00030},
  year={2022}
}

@inproceedings{uniact,
  title={Universal actions for enhanced embodied foundation models},
  author={Zheng, Jinliang and Li, Jianxiong and Liu, Dongxiu and Zheng, Yinan and Wang, Zhihao and Ou, Zhonghong and Liu, Yu and Liu, Jingjing and Zhang, Ya-Qin and Zhan, Xianyuan},
  booktitle={Proceedings of the Computer Vision and Pattern Recognition Conference},
  pages={22508--22519},
  year={2025}
}

@article{motubrain,
  title={MotuBrain: An Advanced World Action Model for Robot Control},
  author={Team, MotuBrain and Xiang, Chendong and Bao, Fan and Liu, Haitian and Tan, Hengkai and Bi, Hongzhe and Li, James and Liu, Jiabao and Pang, Jingrui and Jing, Kiro and others},
  journal={arXiv preprint arXiv:2604.27792},
  year={2026}
}

@article{dit4dit,
  title={Dit4dit: Jointly modeling video dynamics and actions for generalizable robot control},
  author={Ma, Teli and Zheng, Jia and Wang, Zifan and Jiang, Chunli and Cui, Andy and Liang, Junwei and Yang, Shuo},
  journal={arXiv preprint arXiv:2603.10448},
  year={2026}
}

@inproceedings{liv,
  title={Liv: Language-image representations and rewards for robotic control},
  author={Ma, Yecheng Jason and Kumar, Vikash and Zhang, Amy and Bastani, Osbert and Jayaraman, Dinesh},
  booktitle={International Conference on Machine Learning},
  pages={23301--23320},
  year={2023},
  organization={PMLR}
}

@article{world-models,
  title={World models},
  author={Ha, David and Schmidhuber, J{\"u}rgen},
  journal={arXiv preprint arXiv:1803.10122},
  volume={2},
  number={3},
  pages={440},
  year={2018}
}

@article{world-models-survey,
  title={Understanding world or predicting future? a comprehensive survey of world models},
  author={Ding, Jingtao and Zhang, Yunke and Shang, Yu and Zhang, Yuheng and Zong, Zefang and Feng, Jie and Yuan, Yuan and Su, Hongyuan and Li, Nian and Sukiennik, Nicholas and others},
  journal={ACM Computing Surveys},
  volume={58},
  number={3},
  pages={1--38},
  year={2025},
  publisher={ACM New York, NY}
}

@article{jepa,
  title={V-jepa 2: Self-supervised video models enable understanding, prediction and planning},
  author={Assran, Mido and Bardes, Adrien and Fan, David and Garrido, Quentin and Howes, Russell and Muckley, Matthew and Rizvi, Ammar and Roberts, Claire and Sinha, Koustuv and Zholus, Artem and others},
  journal={arXiv preprint arXiv:2506.09985},
  year={2025}
}

@article{remix,
  title={Re-mix: Optimizing data mixtures for large scale imitation learning},
  author={Hejna, Joey and Bhateja, Chethan and Jiang, Yichen and Pertsch, Karl and Sadigh, Dorsa},
  journal={arXiv preprint arXiv:2408.14037},
  year={2024}
}

@article{datamil,
  title={Datamil: Selecting data for robot imitation learning with datamodels},
  author={Dass, Shivin and Khaddaj, Alaa and Engstrom, Logan and Madry, Aleksander and Ilyas, Andrew and Mart{\'\i}n-Mart{\'\i}n, Roberto},
  journal={arXiv preprint arXiv:2505.09603},
  year={2025}
}

@article{rl-token,
  title={RL Token: Bootstrapping Online RL with Vision-Language-Action Models},
  author={Xu, Charles and Springenberg, Jost Tobias and Equi, Michael and Amin, Ali and Esmail, Adnan and Levine, Sergey and Ke, Liyiming},
  journal={arXiv preprint arXiv:2604.23073},
  year={2026}
}

@inproceedings{c51,
  title={A distributional perspective on reinforcement learning},
  author={Bellemare, Marc G and Dabney, Will and Munos, R{\'e}mi},
  booktitle={International conference on machine learning},
  pages={449--458},
  year={2017},
  organization={Pmlr}
}

@inproceedings{dit,
  title={Scalable diffusion models with transformers},
  author={Peebles, William and Xie, Saining},
  booktitle={Proceedings of the IEEE/CVF international conference on computer vision},
  pages={4195--4205},
  year={2023}
}

@InProceedings{value,
  title = 	 {Universal Value Function Approximators},
  author = 	 {Schaul, Tom and Horgan, Daniel and Gregor, Karol and Silver, David},
  booktitle = 	 {Proceedings of the 32nd International Conference on Machine Learning},
  pages = 	 {1312--1320},
  year = 	 {2015},
  editor = 	 {Bach, Francis and Blei, David},
  volume = 	 {37},
  series = 	 {Proceedings of Machine Learning Research},
  address = 	 {Lille, France},
  month = 	 {07--09 Jul},
  publisher =    {PMLR},
  url = 	 {https://proceedings.mlr.press/v37/schaul15.html}
}

@inproceedings{awr-control,
  title={Reinforcement learning by reward-weighted regression for operational space control},
  author={Peters, Jan and Schaal, Stefan},
  booktitle={Proceedings of the 24th international conference on Machine learning},
  pages={745--750},
  year={2007}
}

@article{awr,
  title={Advantage-weighted regression: Simple and scalable off-policy reinforcement learning},
  author={Peng, Xue Bin and Kumar, Aviral and Zhang, Grace and Levine, Sergey},
  journal={arXiv preprint arXiv:1910.00177},
  year={2019}
}

@article{dreamer,
  title={Dream to control: Learning behaviors by latent imagination},
  author={Hafner, Danijar and Lillicrap, Timothy and Ba, Jimmy and Norouzi, Mohammad},
  journal={arXiv preprint arXiv:1912.01603},
  year={2019}
}

@article{alphago,
  title={Mastering the game of Go with deep neural networks and tree search},
  author={Silver, David and Huang, Aja and Maddison, Chris J and Guez, Arthur and Sifre, Laurent and Van Den Driessche, George and Schrittwieser, Julian and Antonoglou, Ioannis and Panneershelvam, Veda and Lanctot, Marc and others},
  journal={nature},
  volume={529},
  number={7587},
  pages={484--489},
  year={2016},
  publisher={Nature Publishing Group UK London}
}

@article{vpn,
  title={Value prediction network},
  author={Oh, Junhyuk and Singh, Satinder and Lee, Honglak},
  journal={Advances in neural information processing systems},
  volume={30},
  year={2017}
}

@article{muzero,
  title={Mastering atari, go, chess and shogi by planning with a learned model},
  author={Schrittwieser, Julian and Antonoglou, Ioannis and Hubert, Thomas and Simonyan, Karen and Sifre, Laurent and Schmitt, Simon and Guez, Arthur and Lockhart, Edward and Hassabis, Demis and Graepel, Thore and others},
  journal={Nature},
  volume={588},
  number={7839},
  pages={604--609},
  year={2020},
  publisher={Nature Publishing Group UK London}
}

@article{pomdp,
  title={Planning and acting in partially observable stochastic domains},
  author={Kaelbling, Leslie Pack and Littman, Michael L and Cassandra, Anthony R},
  journal={Artificial intelligence},
  volume={101},
  number={1-2},
  pages={99--134},
  year={1998},
  publisher={Elsevier}
}

@inproceedings{recurrentQ,
  title={Deep Recurrent Q-Learning for Partially Observable MDPs.},
  author={Hausknecht, Matthew J and Stone, Peter},
  booktitle={AAAI fall symposia},
  volume={45},
  pages={141},
  year={2015}
}

@article{longterm,
  title={Learning long-term dependencies with gradient descent is difficult},
  author={Bengio, Yoshua and Simard, Patrice and Frasconi, Paolo},
  journal={IEEE transactions on neural networks},
  volume={5},
  number={2},
  pages={157--166},
  year={1994},
  publisher={IEEE}
}

@article{lstmRL,
  title={Reinforcement learning with long short-term memory},
  author={Bakker, Bram},
  journal={Advances in neural information processing systems},
  volume={14},
  year={2001}
}

@article{seedance,
  title={Seedance 2.0: Advancing video generation for world complexity},
  author={Seedance, Team and Chen, De and Chen, Liyang and Chen, Xin and Chen, Ying and Chen, Zhuo and Chen, Zhuowei and Cheng, Feng and Cheng, Tianheng and Cheng, Yufeng and others},
  journal={arXiv preprint arXiv:2604.14148},
  year={2026}
}

@article{sora,
  title={Sora: A review on background, technology, limitations, and opportunities of large vision models},
  author={Liu, Yixin and Zhang, Kai and Li, Yuan and Yan, Zhiling and Gao, Chujie and Chen, Ruoxi and Yuan, Zhengqing and Huang, Yue and Sun, Hanchi and Gao, Jianfeng and others},
  journal={arXiv preprint arXiv:2402.17177},
  year={2024}
}

@inproceedings{interplay,
  title={Studying the interplay between the actor and critic representations in reinforcement learning},
  author={Garcin, Samuel and McInroe, Trevor and Castro, Pablo Samuel and Lucas, Christopher and Abel, David and Panangaden, Prakash and Albrecht, Stefano V},
  booktitle={International Conference on Learning Representations},
  volume={2025},
  pages={35590--35616},
  year={2025}
}

@article{flow,
  title={Flow matching for generative modeling},
  author={Lipman, Yaron and Chen, Ricky TQ and Ben-Hamu, Heli and Nickel, Maximilian and Le, Matt},
  journal={arXiv preprint arXiv:2210.02747},
  year={2022}
}

@article{cfg,
  title={Classifier-free diffusion guidance},
  author={Ho, Jonathan and Salimans, Tim},
  journal={arXiv preprint arXiv:2207.12598},
  year={2022}
}

@article{pearson,
  title={VII. Note on regression and inheritance in the case of two parents},
  author={Pearson, Karl},
  journal={proceedings of the royal society of London},
  volume={58},
  number={347-352},
  pages={240--242},
  year={1895},
  publisher={The Royal Society London}
}

@article{droid,
  title={Droid: A large-scale in-the-wild robot manipulation dataset},
  author={Khazatsky, Alexander and Pertsch, Karl and Nair, Suraj and Balakrishna, Ashwin and Dasari, Sudeep and Karamcheti, Siddharth and Nasiriany, Soroush and Srirama, Mohan Kumar and Chen, Lawrence Yunliang and Ellis, Kirsty and others},
  journal={arXiv preprint arXiv:2403.12945},
  year={2024}
}

@article{physiagent,
  title={PhysiAgent: An Embodied Agent Framework in Physical World},
  author={Wang, Zhihao and Li, Jianxiong and Zheng, Jinliang and Zhang, Wencong and Liu, Dongxiu and Zheng, Yinan and Niu, Haoyi and Yu, Junzhi and Zhan, Xianyuan},
  journal={arXiv preprint arXiv:2509.24524},
  year={2025}
}

@article{mem,
  title={Mem: Multi-scale embodied memory for vision language action models},
  author={Torne, Marcel and Pertsch, Karl and Walke, Homer and Vedder, Kyle and Nair, Suraj and Ichter, Brian and Ren, Allen Z and Wang, Haohuan and Tang, Jiaming and Stachowicz, Kyle and others},
  journal={arXiv preprint arXiv:2603.03596},
  year={2026}
}

@article{lbp,
  title={Efficient Robotic Policy Learning via Latent Space Backward Planning},
  author={Liu, Dongxiu and Niu, Haoyi and Wang, Zhihao and Zheng, Jinliang and Zheng, Yinan and Ou, Zhonghong and Hu, Jianming and Li, Jianxiong and Zhan, Xianyuan},
  journal={arXiv preprint arXiv:2505.06861},
  year={2025}
}

@inproceedings{mutual,
  title={Robo-mutual: Robotic multimodal task specification via unimodal learning},
  author={Li, Jianxiong and Wang, Zhihao and Zheng, Jinliang and Zhou, Xiaoai and Wang, Guanming and Song, Guanglu and Liu, Yu and Liu, Jingjing and Zhang, Ya-Qin and Yu, Junzhi and others},
  booktitle={2025 IEEE International Conference on Robotics and Automation (ICRA)},
  pages={4182--4189},
  year={2025},
  organization={IEEE}
}

@article{sop,
  title={SOP: A Scalable Online Post-Training System for Vision-Language-Action Models},
  author={Pan, Mingjie and Feng, Siyuan and Zhang, Qinglin and Li, Xinchen and Song, Jianheng and Qu, Chendi and Wang, Yi and Li, Chuankang and Xiong, Ziyu and Chen, Zhi and others},
  journal={arXiv preprint arXiv:2601.03044},
  year={2026}
}

@article{leworldmodel,
  title={Leworldmodel: Stable end-to-end joint-embedding predictive architecture from pixels},
  author={Maes, Lucas and Lidec, Quentin Le and Scieur, Damien and LeCun, Yann and Balestriero, Randall},
  journal={arXiv preprint arXiv:2603.19312},
  year={2026}
}

@article{rectifiedflow,
  title={Flow straight and fast: Learning to generate and transfer data with rectified flow},
  author={Liu, Xingchao and Gong, Chengyue and Liu, Qiang},
  journal={arXiv preprint arXiv:2209.03003},
  year={2022}
}

@article{flow-tutorial,
  title={Flow matching guide and code},
  author={Lipman, Yaron and Havasi, Marton and Holderrieth, Peter and Shaul, Neta and Le, Matt and Karrer, Brian and Chen, Ricky TQ and Lopez-Paz, David and Ben-Hamu, Heli and Gat, Itai},
  journal={arXiv preprint arXiv:2412.06264},
  year={2024}
}

@book{rl-intro,
  title={Reinforcement learning: An introduction},
  author={Sutton, Richard S and Barto, Andrew G and others},
  volume={1},
  number={1},
  year={1998},
  publisher={MIT press Cambridge}
}

@article{creditsurvey,
  title={A survey of temporal credit assignment in deep reinforcement learning},
  author={Pignatelli, Eduardo and Ferret, Johan and Geist, Matthieu and Mesnard, Thomas and van Hasselt, Hado and Pietquin, Olivier and Toni, Laura},
  journal={arXiv preprint arXiv:2312.01072},
  year={2023}
}

@article{tvt,
  title={Optimizing agent behavior over long time scales by transporting value},
  author={Hung, Chia-Chun and Lillicrap, Timothy and Abramson, Josh and Wu, Yan and Mirza, Mehdi and Carnevale, Federico and Ahuja, Arun and Wayne, Greg},
  journal={Nature communications},
  volume={10},
  number={1},
  pages={5223},
  year={2019},
  publisher={Nature Publishing Group UK London}
}

@inproceedings{gtrxl,
  title={Stabilizing transformers for reinforcement learning},
  author={Parisotto, Emilio and Song, Francis and Rae, Jack and Pascanu, Razvan and Gulcehre, Caglar and Jayakumar, Siddhant and Jaderberg, Max and Kaufman, Raphael Lopez and Clark, Aidan and Noury, Seb and others},
  booktitle={International conference on machine learning},
  pages={7487--7498},
  year={2020},
  organization={PMLR}
}

@inproceedings{rpg,
  title={Solving deep memory POMDPs with recurrent policy gradients},
  author={Wierstra, Daan and Foerster, Alexander and Peters, Jan and Schmidhuber, Juergen},
  booktitle={International conference on artificial neural networks},
  pages={697--706},
  year={2007},
  organization={Springer}
}

@inproceedings{r2d2,
  title={Recurrent experience replay in distributed reinforcement learning},
  author={Kapturowski, Steven and Ostrovski, Georg and Quan, John and Munos, Remi and Dabney, Will},
  booktitle={International conference on learning representations},
  year={2018}
}

@article{gigaworld,
  title={GigaWorld-Policy: An Efficient Action-Centered World--Action Model},
  author={Ye, Angen and Wang, Boyuan and Ni, Chaojun and Huang, Guan and Zhao, Guosheng and Li, Hao and Li, Hengtao and Li, Jie and Lv, Jindi and Liu, Jingyu and others},
  journal={arXiv preprint arXiv:2603.17240},
  year={2026}
}
\end{document}